\documentclass[letterpaper]{article} 
\usepackage{aaai24}  
\usepackage{times}  
\usepackage{helvet}  
\usepackage{courier}  
\usepackage[hyphens]{url}  
\usepackage{graphicx} 
\usepackage{natbib}  
\usepackage{caption} 
\usepackage{algorithm}
\usepackage{algorithmic}

\usepackage{newfloat}
\usepackage{listings}
\DeclareCaptionStyle{ruled}{labelfont=normalfont,labelsep=colon,strut=off} 
\floatstyle{ruled}
\newfloat{listing}{tb}{lst}{}
\floatname{listing}{Listing}
\usepackage{epsfig}
\usepackage{booktabs}
\usepackage{multirow}
\usepackage{amsmath}
\nocopyright 

\title{Better than Random: Reliable NLG Human Evaluation with \\Constrained Active Sampling}
\author{
    Jie Ruan, Xiao Pu, Mingqi Gao, Xiaojun Wan, Yuesheng Zhu
}
\affiliations{
    Peking University\\
    \{ruanjie,puxiao\}@stu.pku.edu.cn, \{gaomingqi,wanxiaojun,zhuys\}@pku.edu.cn
}

\usepackage{bibentry}

\begin{document}

\maketitle

\begin{abstract}
Human evaluation is viewed as a reliable evaluation method for NLG which is expensive and time-consuming. To save labor and costs, researchers usually perform human evaluation on a small subset of data sampled from the whole dataset in practice. However, different selection subsets will lead to different rankings of the systems. 
To give a more correct inter-system ranking and make the gold standard human evaluation more reliable, we propose a Constrained Active Sampling Framework (CASF) for reliable human judgment. 
CASF operates through a Learner, a Systematic Sampler and a Constrained Controller to select representative samples for getting a more correct inter-system ranking.
Experiment results on 137 real NLG evaluation setups with 44 human evaluation metrics across 16 datasets and 5 NLG tasks demonstrate CASF receives 93.18\% top-ranked system recognition accuracy and ranks first or ranks second on 90.91\% of the human metrics with 0.83 overall inter-system ranking Kendall correlation.
Code and data are publicly available online.
\end{abstract}
\begin{figure}[t]
   \centering
    \includegraphics[width=0.499\textwidth,height=0.49\textwidth]{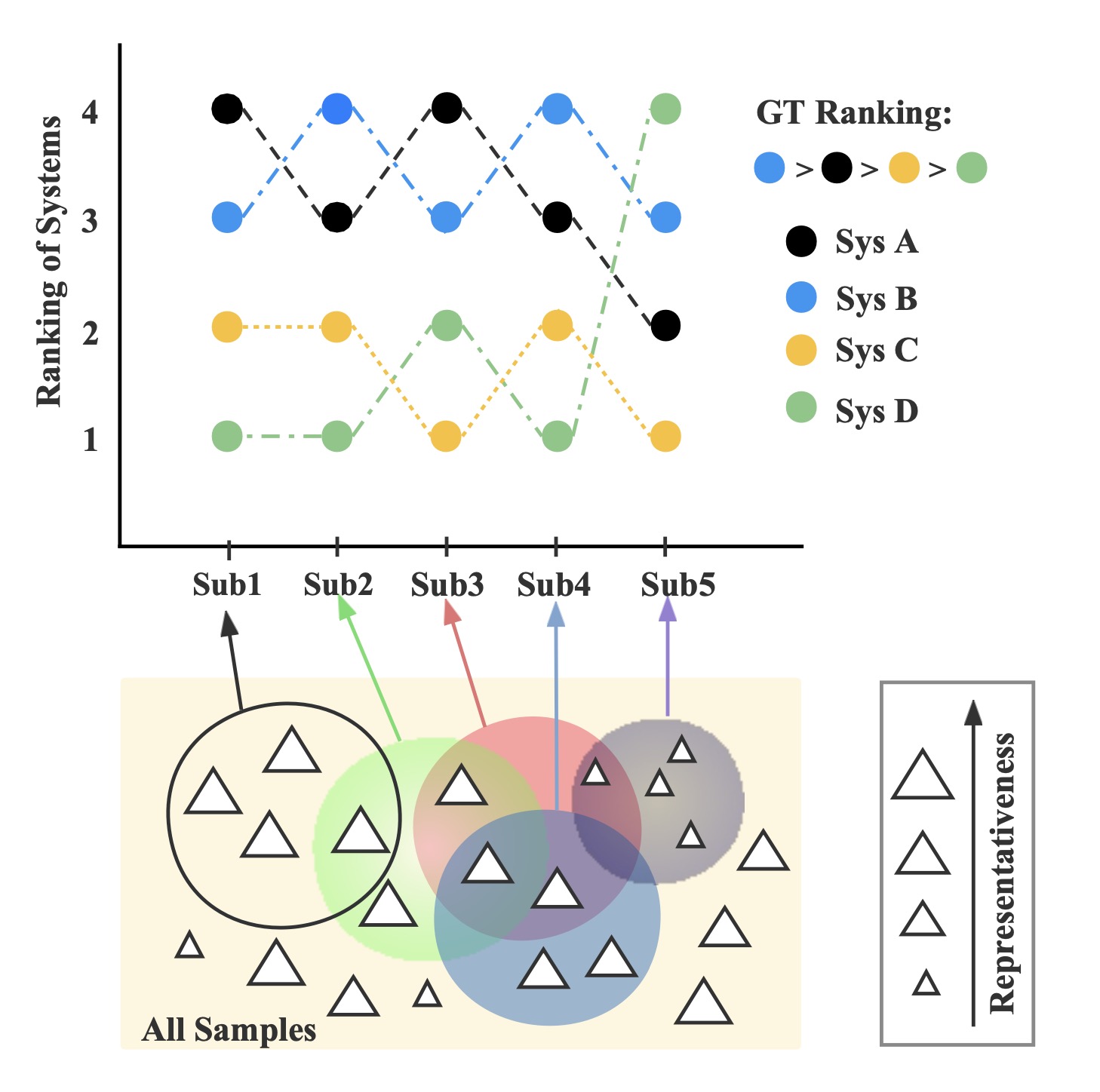}
    \caption{Conducting human evaluations on different sample subsets (Sub) can obtain different inter-system rankings. The lower part shows the same sampling method obtains different subsets at different sampling times. The upper part shows the ranking obtained from the corresponding subsets. ``Sys" represents system and ``GT" represents Ground Truth.}
    \label{fig:motivation}
\end{figure} 

\section{Introduction}

Evaluation of NLG systems remains challenging. The reason is that similar content in text can often be expressed in various ways, and the same output of the NLG system may need to satisfy multiple goals in different aspects \shortcite{howcroft2020twenty,zhou2022deconstructing}. 
Hence, reliable automatic metrics are complex to design \shortcite{novikova2017we,reiter2009investigation}. Human evaluation is generally considered to be a more reliable evaluation way in natural language generation tasks \shortcite{celikyilmaz2020evaluation,gatt2018survey,gkatzia2015snapshot}. However, human judgment is viewed as expensive, time-consuming, and lacks standardized evaluation procedures \shortcite{howcroft2020twenty,celikyilmaz2020evaluation,mohankumar2022active}.

To save labor and costs, human evaluation is usually performed on a small subset sampled from the dataset in practice. Researchers compare the average scores of the systems on this subset to obtain a ranking between the systems.
However, different sample subsets will lead to different rankings of the systems. 
We re-evaluated 137 real NLG evaluation setups on 44 human metrics across 16 datasets and 5 NLG tasks. Results show that 87.5\% of datasets have different inter-system rankings across 5 times of random sampling.
Since research is driven by evaluation, focusing on the final ranking of systems, it is vital to design a more reliable evaluation method to obtain the correct inter-system ranking.

We randomly select 1404 papers from ACL, EMNLP and COLING in the past 2 years and find that 270 papers select a subset of the dataset for manual evaluation to save labor and cost (details are in the Survey section of the Appendix). The survey results show that random sampling is the most vital sampling method, accounting for 60.7\%, and the rest 39.3\% of the papers do not mention the sampling method they used. 
Random sampling is widely used in human evaluation sampling for its simplicity.
However, random sampling can be risky \shortcite{bethard2022we}.
On the one hand, random sampling can lead to clustered selection, 
a phenomenon in which randomly selected samples are uncommonly close together in a population (as shown in the black and purple circle in Figure \ref{fig:motivation}).
On the other hand, random sampling may have the risk of data manipulation. Researchers can choose samples at will or conduct multiple random sampling to select a favorite subset, which will lead to unfair evaluation results.
Since different sampling subsets may result in different inter-system rankings in human judgment, it is difficult to reliably select the best system.
We urgently need a better sampling method to deliver reliable human evaluation with low labor and cost.

In this paper, we focus on improving the reliability of the gold standard human evaluation with limited cost and time used for human annotation. 
Specifically, we explore the problem of clustered selection and data manipulation for manual evaluation sampling and propose a Constrained Active Sampling Framework (CASF) for reliable human judgment. 
The proposed CASF consists of a Learner, a Systematic Sampler and a Constrained Controller. CASF obtains a representative subset of samples in multiple sampling phases.
In each sampling phase, the Learner predicts the quality score for samples and feeds the quality score of each sample to the Systematic Sampler. Then, the Systematic Sampler and the Constrained Controller work together to select representative samples with lower redundancy for the sampling phase. Samples collected in each phase are not duplicates of those collected in previous phases, and will be directly subjected to human evaluation, and the newly labeled ones will also be used to update the Learner.


The main contributions are as follows: 
1) We investigate and experimentally analyze the sampling problem for the gold standard human evaluation in natural language generation. 
2) We propose a Constrained Active Sampling Framework (CASF) for the sampling problem in manual evaluation. The proposed CASF can solve the problem of clustered selection and data manipulation for human evaluation sampling.
3) We re-evaluate 137 real NLG evaluation setups on 44 human evaluation metrics across 16 datasets and 5 NLG tasks. Experiment results demonstrate the proposed method ranks first or ranks second on 90.91\% of the human metrics and receives 93.18\% top-ranked system recognition accuracy.
To ease the adoption of reliable sampling, we release a constrained active sampling tool. We strongly recommend using CASF to sample test instances for human evaluation.
Our tool, code and data are publicly available online.\footnote{https://github.com/EnablerRx/CASF \label{code}}

\section{Methodology}

\subsection{Problem Statement}
The goal of sampling in human evaluation is to select a subset with the intention of estimating the inter-system ranking of the whole sample population. 
Ideally, the obtained subset should cover more representative samples of the population. A good sampling method will result in a more correct inter-system ranking calculated through the sampling subset.

The general evaluation sampling problem is as follows. Given a data set $D = \{(x_i,\mathcal{Y}_i,\mathcal{Q}_i)\}_{i=1}^{N}$ where $N$ is the size of the whole sample population, $x_i$ represents a data input, $\mathcal{Y}_i$ is the corresponding set of generated outputs, $\mathcal{Q}_i$ is the corresponding set of human score vectors. 
The generated output set $\mathcal{Y}_i$ consists of $M$ system outputs and is denoted as $\mathcal{Y}_i = \{ y_{i1},...,y_{ij} \}_{j=1}^{M}$, 
\begin{figure}[t]
    \centering
    \includegraphics[scale=0.43]{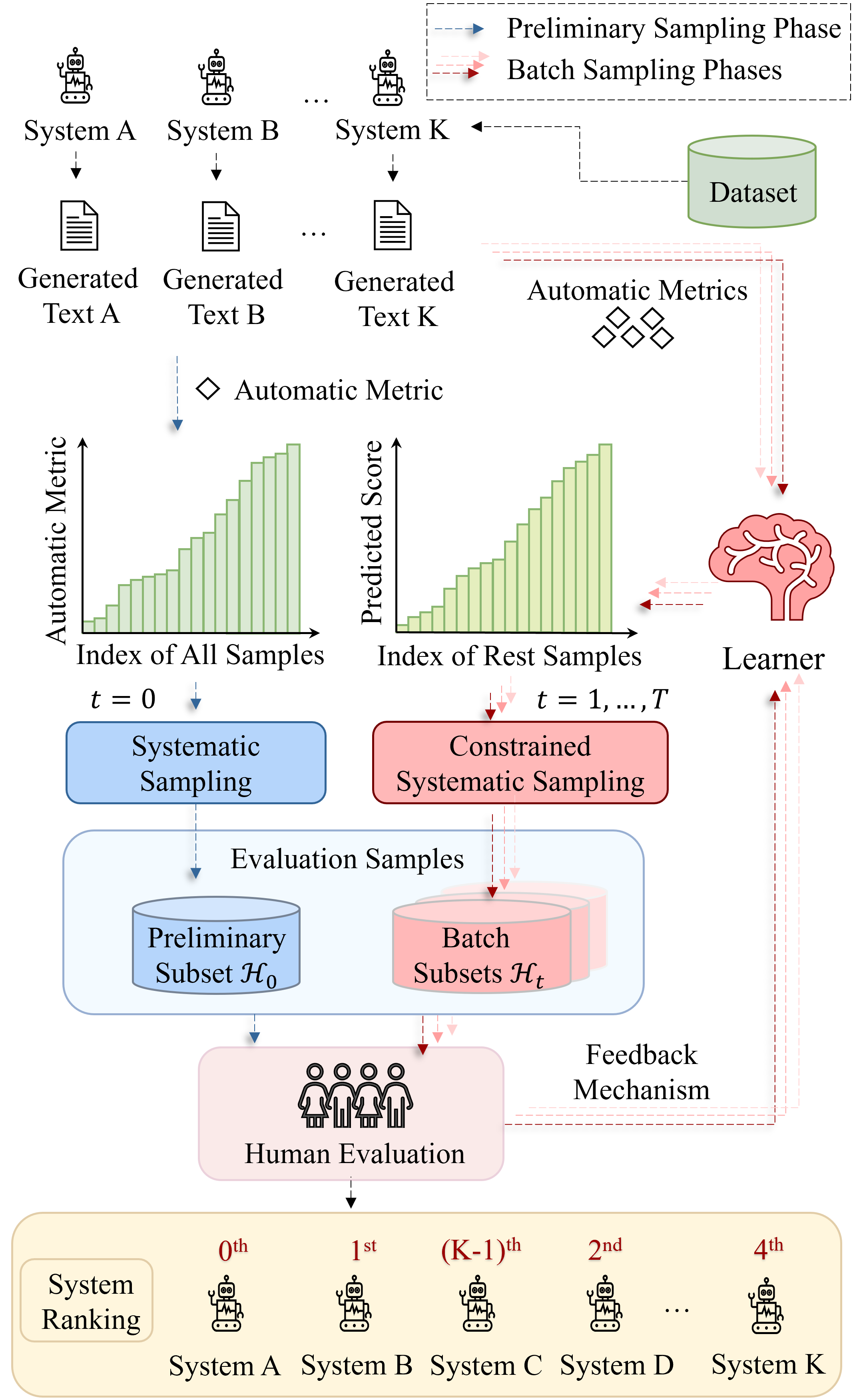}
    \caption{\centering{Constrained Active Sampling Framework}} 
    \label{fig:framework}
\end{figure}
where $y_{ij}$ represents the $j$-th system generated output of the $i$-th sample. 
The human score vector set $\mathcal{Q}_i$ consists of the corresponding human score vector for each system output and is denoted as $\mathcal{Q}_i = \{ \mathbf{q}_{i1},...,\mathbf{q}_{ij}\}_{j=1}^{M}$. Since human evaluation is usually carried out in multiple aspects, we use a vector to represent human evaluation results from multiple aspects for each system.  
Each human score vector $\mathbf{q}_{ij}$ consists of $K$ human annotation metrics from different aspects and is denoted as $\mathbf{q}_{ij} = (q_{ij1},...,q_{ijK})$. Eventually there will be separate inter-system ranking on each aspect.
Let $\mathcal{H}$ represent the final sample subset. Function $\psi$ calculates the mean scores of each system in the sample set for each human evaluation aspect and gives the ranking among systems. $\mathcal{P}$ calculates the similarity between two inter-system rankings.
The overall objective of sampling and constraint is as follows:
\begin{equation*}
\begin{aligned} 
&\text{minimize} \quad -\mathcal{P}[ \psi(\mathcal{H}), \psi(D)],\\
&\text{subject to} \quad |\mathcal{H}|=r\times N,
\end{aligned}
\end{equation*}
where $r$ is the sampling rate, $|.|$ refers to the cardinality of a sample set and $\psi$ first calculates the average human scores in each aspect of each system in the sample set, and then gives the inter-system ranking of each human indicator according to the mean score of each system.

\subsection{Sample Representativeness}
Taking representative samples allows for a more complete evaluation of the overall performance of the system.
Inspired by the theoretical model of summarization \cite{peyrard2019simple}, the \emph{Representativeness} of samples can be measured in two aspects, including \emph{Quality Diversity} and \emph{Redundancy}.
\emph{Quality Diversity} represents the diversity of sample quality levels, that is, the sampled subset should contain samples of various quality levels.
Evaluation on qualitatively diverse subsets of samples allows the system to better reflect the performance of all samples.
Quality is the average quality of generated outputs of the sample. More comprehensive coverage of samples of different qualities will result in a better \emph{Quality Diversity}.
\emph{Redundancy} indicates the degree of similarity or duplication among the generated outputs of samples.

\subsection{Constrained Active Sampling Framework}
\subsubsection{Overall Framework}
The proposed Constrained Active Sampling Framework aims to select representative samples for human evaluation in multiple phases to get a more correct inter-system ranking.
The proposed CASF operates through a Learner, a Systematic Sampler and a Constrained Controller. 
The goal of the Learner is to predict the quality of samples and give a ranking of sample quality by a regressor. 
The Systematic Sampler divides samples into multiple buckets according to the sample quality ranking given by the Learner.
The Constrained Controller controls the \emph{Redundancy} of samples and selects a final sample from each bucket given by the Systematic Sampler.

The proposed Constrained Active Sampling Framework is shown in Figure \ref{fig:framework}.
There are several sampling phases denoted by $t$, an preliminary sampling phase $t = 0$ (the left branch in Figure \ref{fig:framework}) and $T$ batch active sampling phases $t = 1,...,T$ (the right branch in Figure \ref{fig:framework}).
In the preliminary sampling phase, alternate quality scores for all samples are calculated through an automated metric, as the Learner is not ready to use yet. The Systematic Sampler, then, selects a small preliminary subset of samples $\mathcal{H}_0$ as part of the final sample subset $\mathcal{H}$ according to the given quality ranking.
The selected samples are then evaluated by human beings.
In the current batch active sampling phase, samples selected in all previous phases together with the corresponding human scores, then, are fed to the regressor of the learner, and the regressor of the learner is updated and applied to predict the quality of the rest samples with the sample's scores over various automatic metrics as features.
After that, the Systematic Sampler and Constrained Controller work together to choose batch subset $\mathcal{H}_{t}$ from the rest samples for the $t$-th batch active sampling phase as part of the final samples. 
Then, the samples selected in the $t$-th phase are subjected to human evaluation for use in the subsequent sampling phases. 
The final sample set $\mathcal{H}$ consists of batch subsets from each phase $\mathcal{H}_t$.
We conduct experiments to explore the determination of the number of phases and the sampling ratio of each phase in the Phases and Associated Sampling Ratios section.

\subsubsection{Learner and Sample Quality}
Estimating the quality of the samples is a vital step in CASF. Since the quality of samples is difficult to define and calculate directly, we propose a Learner to predict the human scores as the quality scores for the rest samples for selection at each phase $t$ (except the preliminary phase). As various automatic metrics can measure the characteristics of samples in different aspects and are easy to calculate with lower cost, we use scores of automatic metrics as features to predict the quality of samples.

Note that in the preliminary phase, the quality of samples is simply estimated by an automatic metric. In each of the batch active sampling phases, the Learner receives feedback from human annotators and update its parameters. After that, it utilizes the scores of  automatic metrics to predict the quality score for each sample. The Learner will then provide the quality ranking $\{ p_t(i)\}_{i=1}^{N-|\mathcal{H}|}$ of samples at each batch $t$, where $i$ is the sample index and the number of the rest samples for selection in each phase is $N-|\mathcal{H}|$.

The main objective of the Learner $g$ is to map $x_i$ to the corresponding human score vector set $\mathcal{Q}_i$. 
Since there are multiple elements in $\mathcal{Q}_i$, we standardize scores for each human evaluation aspect and use the sum of each element in $\mathcal{Q}_i$, which is the sum of human scores for all aspects of all NLG systems under sample $x_i$, to represent $\mathcal{Q}_i$. The objective is to minimize the following loss function:
\begin{equation*}
\underset{\theta_t}{\operatorname{argmin}}
\sum_{i=1}^{|\mathcal{H}|}\mathcal{L}\left( g(x_i ; \theta_{t}), \sum_{j=1}^{M} \sum_{k=1}^{K} q_{ijk} \right),
\end{equation*}
where $|\mathcal{H}|$ is the number of samples selected in the final subset and $\theta_t$ is the parameter of Learner $g$ in the $t$-th phase.
The predicted quality scores $\{ s_t(i)\}_{i=1}^{N-|\mathcal{H}|}$ for the rest samples at each phase $t$ are calculated as follows:
\begin{equation*}
\{s_t(i)\}_{i=1}^{N-|\mathcal{H}|} = \{ g(x_i ; \theta_{t} )\}_{i=1}^{N-|\mathcal{H}|} .    
\end{equation*}
Specifically, the Learner first calculates the results of each automatic metric based on the output of each NLG system from the input sample. Then, the automatic metric results under each NLG system will be fed as features into the Learner's regressor. 
Eight popular NLG metrics are chosen as the automatic metrics set (details are in the Automatic Metric for Preliminary Phase section) of CASF.
Due to the small number of samples and features mainly containing automatic metrics' scores, we explore several popular learning methods and recommend choosing Gradient Boosting Decision Tree (GBDT) \cite{friedman2001greedy} as the regressor of the Learner.
Full experimental results are in the Learner Selection section of 
Appendix. The loss function is the least squares method \shortcite{abdi2007method}, which is commonly used in GBDT.

\subsubsection{Systematic Sampler}
Systematic sampling has advantage of eliminating clustered selection problem and can reduce the risk of favoritism, which meets our motivation. Therefore, we adopt the systematic sampling method \cite{yates1948systematic} sorted by relevant signs as the sampling core of CASF.
The Systematic Sampler selects representative initial samples and candidate samples according to the quality ranking of samples.
Specifically, the Systematic Sampler first divides the $N_t=N-|\mathcal{H}|$ samples for the $t$-th phase  into $n_{t}$ buckets according to the given quality ranking $\{ p_{t}(i)\}_{i=1}^{N-|\mathcal{H}|}$. $n_{t}$ is the number of samples to be selected at the $t$-th phase.
Samples with quality ranking $p_{t} \in [e \times \lfloor \frac{N_t}{n_{t}} \rfloor, (e+1) \times \lfloor \frac{N_t}{n_{t}} \rfloor)$ are divided into the same bucket, where $e = 0,1,...,n_{t}$. The samples with quality rank $ p_{t} = e \times \lfloor \frac{N_t}{n_{t}} \rfloor $ are selected as the \textbf{initial selection samples}. And the rest samples in each bucket are \textbf{candidate samples}.

\subsubsection{Constrained Controller}
\begin{figure}[]
    \centering
    \includegraphics[width=0.45\textwidth,height=0.35\textwidth]{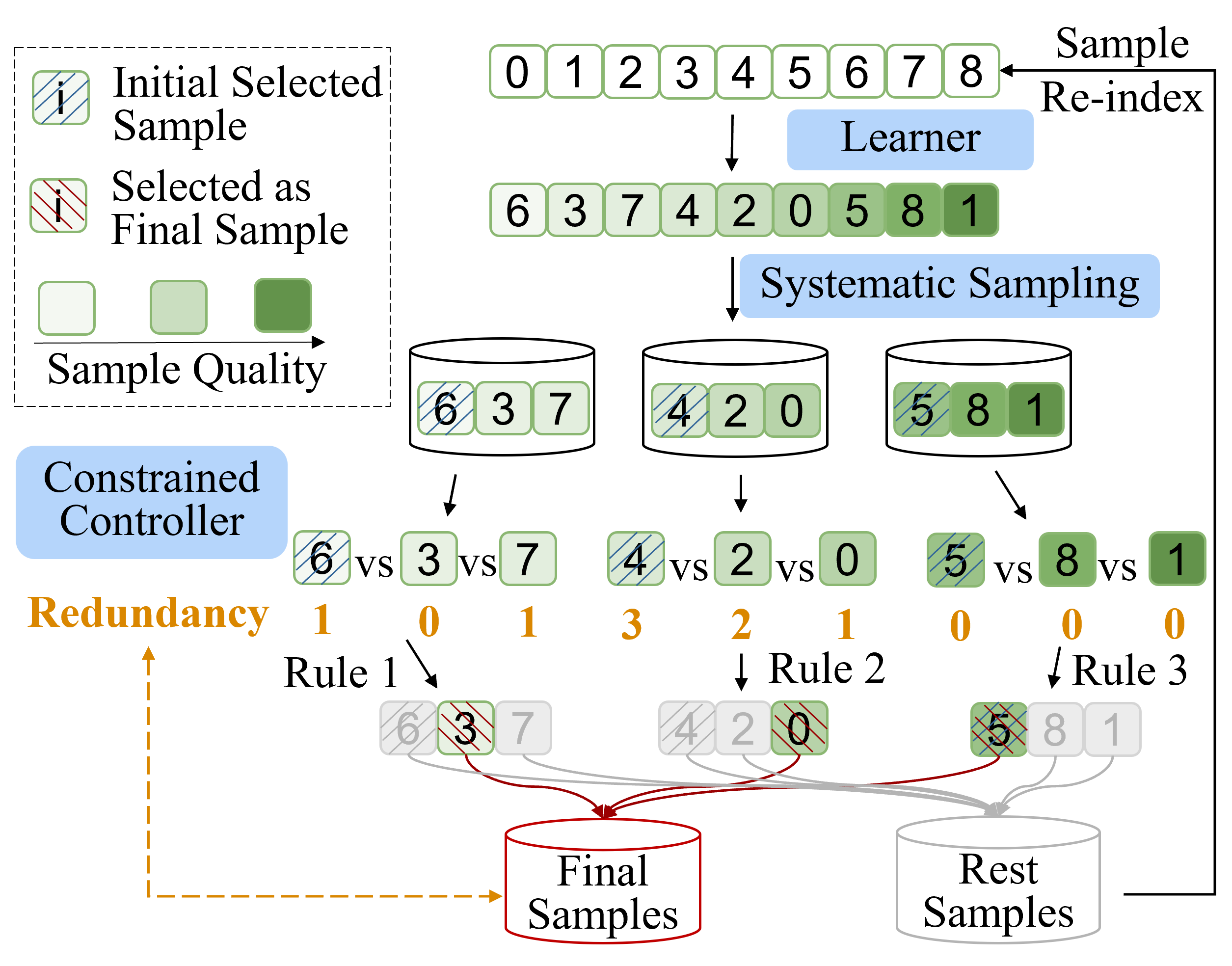}
    \caption{Example of systematic sampler and constrained controller cooperating to select final samples}
    \label{fig:ConstrainedController}
\end{figure}

The proposed Constrained Controller controls the \emph{Redundancy} of samples and selects one sample from each of the buckets divided by the Systematic Sampler to form a final sample subset (as shown in Figure \ref{fig:ConstrainedController}). Since the Systematic Sampler selects initial samples at a regular interval, which makes the distribution of the initial subset align closely with the overall distribution, we aim to preserve the original sampling intervals as much as possible while controlling the Redundancy to maintain the representativeness of the sample subset.
 
Specifically, we define objective function $\operatorname{Obj}$ as the quality ranking distance between the current sample $x_i$ and the initial selection sample in each bucket. We also define violation function $\operatorname{Vio}$ to calculate the \emph{Redundancy} between the current sample $x_i$ and the final samples. 
Since the bi-gram similarity \cite{kondrak2005n} is regarded as a simple and effective method to calculate the redundancy between texts, we calculate the \emph{Redundancy} by calculating the bi-gram similarity between the outputs generated for the sample and that for the final samples. 
A sample $x_i$ is called feasible if $\operatorname{Vio}(x_i) = 0$, which means it is not redundant with the selected final samples. Otherwise, $x_i$ is infeasible.

The Constrained Controller is summarized into 3 rules:
\begin{equation*}
\left \{
\begin{array}{ll}
rule\ 1 : \ x_i \prec x_j ,\ \ \text{if}&x_i \text { is infeasible, } x_j \text { is feasible};\\
rule\ 2 : \ x_i \prec x_j ,\ \ \text {if} 
&\operatorname{Vio}\left(x_i\right)>\operatorname{Vio}\left(x_j\right),\\& x_i \text { is infeasible, } x_j \text { is infeasible};\\
rule\ 3 : \ x_i \prec x_j ,\ \ \text {if}& \operatorname{Obj}\left(x_i\right)>\operatorname{Obj}\left(x_j\right), \\& x_i \text { is feasible, } x_j \text { is feasible},
\end{array}
\right .
\end{equation*}
where $x_i$ is the $i$-th sample and $x_i \prec x_j$ means $x_j$ is a better choice.
$rule\ 1$ means the Constrained Controller tends to select samples that are not redundant. $rule\ 2$ represents that if two samples are both redundant with the final samples, the Constrained Controller tends to select samples with less redundancy. $rule\ 3$ demonstrates that if two samples are both not redundant with the final samples, the Constrained Controller tends to select samples with ranks as close as possible to those of the initial selection samples.

In Figure \ref{fig:ConstrainedController}, the rest samples for selection are first re-indexed, and then re-ordered according to Learner's predicted quality score. The system sampler divides samples into three buckets based on quality ranking and marks initial selection sample for each bucket.
In the first bucket, only sample 3 is feasible, that is, sample 3 is not redundant with existing final samples. Thus, Sample 3 is selected as the final sample according to $rule\ 1$.
In the second bucket, none of the three samples is feasible, so sample 0 with the smallest redundancy is selected as the final sample according to $rule\ 2$.
In the third bucket, all samples are feasible, and sample 5 is the initial selection sample and it is selected by default or according to $rule\ 3$.

\begin{table*}[htbp]
\small
\centering
\begin{tabular}{ccllllllllllll}
\toprule
Dataset & HE Metric & \multicolumn{1}{c}{R 1} & \multicolumn{1}{c}{R 2} & \multicolumn{1}{c}{R 3} & \multicolumn{1}{c}{\begin{tabular}[c]{@{}c@{}}R\\       Mean\end{tabular}} & \multicolumn{1}{c}{H 1} & \multicolumn{1}{c}{H 2} & \multicolumn{1}{c}{H 3} & \multicolumn{1}{c}{\begin{tabular}[c]{@{}c@{}}H\\       Mean\end{tabular}} & \multicolumn{1}{c}{8M} & \multicolumn{1}{c}{SM} & \multicolumn{1}{c}{OL} & \multicolumn{1}{c}{\begin{tabular}[c]{@{}c@{}}CASF\\       (ours)\end{tabular}} \\ \hline\hline
\multirow{4}{*}{SummEval} & coherence & 0.85 & 0.65 & 0.33 & 0.61 & 0.70 & 0.82 & 0.92 & 0.81 & 0.42 & 0.42 & {\underline{0.87}} & \textbf{0.95} \\
 & consistency & 0.25 & 0.48 & 0.43 & 0.39 & 0.68 & 0.02 & 0.65 & 0.45 & 0.30  & 0.17 & \textbf{0.53} & \textbf{0.53} \\
& fluency & 0.40 & 0.35 & 0.52 & {\underline{0.42}} & 0.45 & 0.45 & 0.30 & 0.40  & 0.35  & 0.37 & \textbf{0.52} & 0.33 \\
& relevance & 0.72 & 0.60 & 0.68 & {\underline{0.67}} & 0.65 & 0.43 & 0.72 & 0.60 & 0.40 & 0.60 & 0.45 & \textbf{0.82} \\\hline
REALSumm & litepyramid & 0.39 & 0.54 & 0.44 & 0.46 & 0.36 & 0.38 & 0.44 & 0.39 & 0.33 & 0.37 & \textbf{0.54} & \textbf{0.54} \\\hline
\multirow{4}{*}{NeR18} & coherence & 1.00 & 1.00 & 0.43 & 0.81 & 0.90 & 0.90 & 0.90 & 0.90 &  \textbf{1.00}  & \textbf{1.00} & \textbf{1.00} & \textbf{1.00} \\
& fluency & 0.52 & 1.00 & 1.00 & 0.84 & 1.00 & 0.52 & 0.90 & 0.81 & \textbf{1.00} & \textbf{1.00} & \textbf{1.00} & \textbf{1.00} \\
& informativeness & 1.00 & 1.00 & 1.00 & \textbf{1.00} & 0.71 & 1.00 & 0.90 & 0.87 & 0.71 & \textbf{1.00} & \textbf{1.00} & \textbf{1.00} \\
& relevance & 1.00 & 0.52 & 1.00 & 0.84 & 0.90 & 0.90 & 0.90 & 0.90 & \textbf{1.00} & \textbf{1.00} & \textbf{1.00} & \textbf{1.00} \\\hline
\multirow{4}{*}{DialSumm} & consistency & 0.74 & 0.72 & 0.49 & 0.65 & 0.74 & 0.64 & 0.62 & {\underline{0.67}} & 0.59 & 0.56 & 0.54 & \textbf{0.77} \\
 & relevance & 0.69 & 0.46 & 0.64 & 0.60 & 0.64 & 0.69 & 0.54 & {\underline{0.62}} & 0.23 & 0.44 & 0.59 & \textbf{0.72} \\
 & fluency & 0.59 & 0.56 & 0.59 & 0.58 & 0.38 & 0.56 & 0.51 & 0.49 & 0.15  & 0.49 & \textbf{0.64} & {\underline{0.62}} \\
 & coherence & 0.67 & 0.80 & 0.74 & 0.74 & 0.74 & 0.80 & 0.59 & 0.71 & 0.59 & 0.67 & {\underline{0.82}} & \textbf{0.90} \\\hline
\multirow{4}{*}{OpenAI 1} & accuracy & 0.80 & 0.00 & 1.00 & 0.60 & 0.80 & 1.00 & 0.80 & {\underline{0.87}} & 0.80 & 0.00 & 0.00 & \textbf{1.00} \\
 & coherence & 0.40 & 0.80 & 0.00 & 0.40 & 0.80 & 0.20 & 0.80 & {0.60}  & \textbf{0.80} & 0.40 & 0.20 & \textbf{0.80} \\
 & coverage & 1.00 & 1.00 & 1.00 & \textbf{1.00} & 0.80 & 0.80 & 0.80 & 0.80 & 0.80 & \textbf{1.00} & 0.80 & 0.80 \\
 & overall & 0.80 & 1.00 & 1.00 & 0.93 & 0.80 & 1.00 & 0.80 & 0.87 & 0.80 & \textbf{1.00} & 0.80 & \textbf{1.00} \\\hline
\multirow{4}{*}{OpenAI 2} & accuracy & 0.71 & 0.43 & 1.00 & 0.71 & 0.62 & 0.71 & 0.81 & 0.71 & \textbf{1.00} & 0.52 & 0.14 & {\underline{0.90}} \\
 & coherence & 0.24 & 0.52 & 0.33 & 0.37 & -0.14 & 0.24 & 0.43 & 0.17 & 0.24 & \textbf{0.52} & 0.24 & {\underline{0.43}} \\
 & coverage & 1.00 & 0.71 & 0.90 & 0.87 & 1.00 & 0.90 & 1.00 & 0.97 & \textbf{1.00} & \textbf{1.00} & \textbf{1.00} & \textbf{1.00} \\
 & overall & 0.90 & 0.71 & 1.00 & 0.87 & 0.62 & 1.00 & 0.90 & 0.84 & {\underline{0.90}}  & {\underline{0.90}} & {\underline{0.90}} & \textbf{1.00} \\\hline
\multirow{4}{*}{OpenAI 3} & accuracy & 0.73 & 0.82 & 0.82 & 0.79 & 0.87 & 0.78 & 0.82 & {\underline{0.82}} & 0.73 & 0.69 & 0.78 & \textbf{0.87} \\
 & coherence & 0.51 & 0.33 & 0.56 & 0.47 & 0.42 & 0.51 & 0.56 & 0.50 & 0.56 & 0.20 & \textbf{0.60} & \textbf{0.60} \\
 & coverage & 0.38 & 0.38 & 0.87 & 0.54 & 0.51 & 0.87 & 0.51 & 0.63 & \textbf{1.00}& \textbf{1.00} & 0.42 & {0.87} \\
 & overall & 0.87 & 0.51 & 1.00 & {0.79} & 1.00 & 0.73 & 0.51 & 0.75 & \textbf{1.00} & 0.38 & 0.47 & \textbf{1.00} \\\hline
\multirow{4}{*}{OpenAI 4} & accuracy & 1.00 & 0.33 & 1.00 & 0.78 & 1.00 & 0.33 & 0.33 & 0.56 & 0.33 & \textbf{1.00} & 0.33 & \textbf{1.00} \\
 & coherence & 1.00 & 1.00 & 1.00 & \textbf{1.00} & 1.00 & 1.00 & 1.00 & \textbf{1.00} & \textbf{1.00} & \textbf{1.00} & 0.33 & \textbf{1.00} \\
 & coverage & 0.33 & 1.00 & 1.00 & 0.78 & 0.33 & 1.00 & 1.00 & 0.78 & \textbf{1.00} & \textbf{1.00} & \textbf{1.00} & \textbf{1.00} \\
 & overall & 0.33 & 1.00 & 1.00 & 0.78 & 0.33 & 1.00 & 1.00 & 0.78 & \textbf{1.00} & \textbf{1.00} & \textbf{1.00} & \textbf{1.00} \\ \hline\hline
\multirow{2}{*}{newstest 1} & MQM & 0.14 & 0.14 & 0.14 & {\underline{0.14}} & 0.33 & 0.14 & -0.05 &{\underline{0.14}}& {\underline{0.14}} & \textbf{0.33} & {\underline{0.14}} & {\underline{0.14}} \\
 & pSQM & 0.81 & 0.90 & 0.90 & 0.87 & 0.81 & 0.90 & 0.90 & 0.87 & \textbf{1.00}& { 0.90} & { 0.90} & \textbf{1.00} \\\hline
\multirow{2}{*}{newstest 2} & MQM & 0.79 & 0.93 & 0.71 & 0.81 & 0.64 & 0.86 & 0.71 & 0.74 &0.14 & \textbf{0.93} & 0.86 & \textbf{0.93} \\
 & pSQM & 0.43 & 0.36 & 0.79 & 0.52 & 0.29 & 0.86 & 0.43 & 0.52 & 0.36 & \textbf{0.93} & {\underline{0.79}} & {\underline{0.79}} \\\hline
newstest 3 & MQM & 0.00 & -0.13 & -0.05 & -0.06 & -0.05 & -0.03 & -0.05 & -0.04 & \textbf{0.46} & {\underline{0.13}} & 0.00 & {0.03} \\\hline\hline
\multirow{6}{*}{Persona} & Understandable & 0.33 & -1.00 & 0.33 & -0.11 & -1.00 & 0.33 & 0.33 & -0.11 & \textbf{0.33} & \textbf{0.33} & \textbf{0.33} & \textbf{0.33} \\
 & Natural & 0.33 & -1.00 & 1.00 & 0.11 & 1.00 & -1.00 & 0.33 & 0.11 & {\underline{0.33}}& {\underline{0.33}} & {\underline{0.33}} & \textbf{1.00} \\
 & \multicolumn{1}{c}{\begin{tabular}[c]{@{}c@{}}Maintains\\       Context\end{tabular}} & 1.00 & 1.00 & 1.00 & \textbf{1.00} & 1.00 & 1.00 & 1.00 & \textbf{1.00} & -1.00 & \textbf{1.00} & \textbf{1.00} & \textbf{1.00} \\
 & Interesting & 1.00 & 1.00 & 1.00 & \textbf{1.00} & 1.00 & 0.33 & 1.00 & 0.78 & \textbf{1.00}& \textbf{1.00} & \textbf{1.00} & \textbf{1.00} \\
 & Uses Knowledge & 1.00 & 1.00 & 1.00 & \textbf{1.00} & -1.00 & 1.00 & 1.00 & 0.33 & \textbf{1.00}& \textbf{1.00} & \textbf{1.00} & \textbf{1.00} \\
  & Overall Quality & 1.00 & 1.00 & 1.00 & \textbf{1.00} & 1.00 & 1.00 & 1.00 & \textbf{1.00} & \textbf{1.00} & \textbf{1.00} & \textbf{1.00} & \textbf{1.00} \\\hline\hline
 MANS-ROC & overall & 1.00 & 1.00 & 1.00 & \textbf{1.00} & 1.00 & 1.00 & 1.00 & \textbf{1.00}& \textbf{1.00} & \textbf{1.00} & \textbf{1.00} & \textbf{1.00} \\\hline
MANS-WP & overall & 1.00 & 0.80 & 0.80 & 0.87 & 0.80 & 1.00 & 1.00 & 0.93 & \textbf{1.00}& \textbf{1.00} & \textbf{1.00} & \textbf{1.00} \\\hline\hline
THUMB & overall & 1.00 & 0.80 & 1.00 & 0.93 & 1.00 & 1.00 & 1.00 & \textbf{1.00} & \textbf{1.00} & \textbf{1.00} & \textbf{1.00} & \textbf{1.00} \\\hline
VATEX & consistency & 0.60 & 1.00 & 0.60 & 0.73 & 0.60 & 1.00 & 1.00 & 0.87 & \textbf{1.00} & \textbf{1.00} & \textbf{1.00} & \textbf{1.00} \\\hline\hline
\multicolumn{2}{c}{Overall Performance} & 0.69 & 0.61 & 0.75 & 0.68 & 0.61 & 0.67 &0.65 & 0.72 & 0.67 & {\underline{0.72}} & 0.68 & \textbf{0.83}\\ \bottomrule
\end{tabular}
\caption{Kendall's Tau of methods on 16 datasets across 5 NLG tasks. 'HE Metric' indicates different human evaluation aspects in a dataset. 
{Bold number} indicates that the method has the best performance among all methods under the corresponding aspect. {Underlined number} indicates the method ranks second.
}
\label{tab:result}
\end{table*}
\section{Experimental Setup}

\subsection{Tasks and Datasets } 
We conduct experiments on 44 human  metrics across 16 datasets spanning 5 tasks.
A total of 137 NLG systems are involved.
Details of the datasets, preprocessing and the validation set for hyper-parameters selection are in the Tasks and Dataset section of Appendix. The datasets are: 
\textbf{Summarization (SUM):} We utilize 8 human evaluation datasets of the model generated summarization, which are SummEval \shortcite{fabbri2021summeval}, REALSumm \shortcite{bhandari2020re}, Newsroom (NeR18) \shortcite{grusky2018newsroom}, DialSummEval (DialSumm) \shortcite{gao-wan-2022-dialsummeval} and OpenAI-axis1 (OpenAI 1) \shortcite{stiennon2020learning,volske2017tl}, OpenAI-axis2  (OpenAI 2) , OpenAI-CNN/DM1  (OpenAI 3) , and OpenAI-CNN/DM3  (OpenAI 4) .
\textbf{Machine Translation (MT):} We use 3 datasets collected from WMT news translation tasks \shortcite{freitag2021experts} \emph{viz.} newstest2020 en-de (newstest 1),
newstest2020 cn-en (newstest 2) and newstest2021 cn-en (newstest 3).
\textbf{Dialogue Generation (DGen):} We utilize a human annotation dataset of machine-generated dialogues released with the Persona Chat (Persona) \cite{mehri2020usr} dataset.
\textbf{Story Generation (SGen):} We use two manual evaluation datasets for story generation namely MANS-ROC \cite{guan2021openmeva} and MANS-WP \cite{guan2021openmeva}.
\textbf{Multi-Modal Generation (MMGen):} We use two existing human evaluation datasets namely THUMB-MSCOCO (THUMB) \cite{kasai2022thumb} and VATEX-EVAL (VATEX) \cite{DBLP:conf/cvpr/ShiYXYLHZ22}.

\subsection{Evaluation Metric}
We select a subset of each dataset and then compute the results for all the human metrics in various aspects.
We measure the efficacy of sampling method by computing
rankings of candidate models on the subset and their Kendall’s Tau correlation \shortcite{kendall1938new} with rankings obtained on the full dataset. 
We refer to Kendall’s  treatment \shortcite{kendall1945treatment} to handle ties.

\subsection{Comparison of Methods}
The comparison methods are selected based on the survey of evaluation sampling methods in 1404 papers where Random and Heuristic are the main sampling methods for NLG human evaluation. We also include some ablation methods.
The comparison methods are: 
\textbf{Random Sampling (R)} randomly sample the dataset and is performed 3 times \shortcite{wan2008collabrank,bhatnagar2022chia,wan2007collabsum} to reflect real sampling scenarios. Results of each time and the average result are recorded.
\textbf{Heuristic Sampling (H)} \shortcite{varshney2022ildae} first sorts the samples according to the average length of the generated sentences. Then, Heuristic randomly collects a small number of samples with extreme sentence length and a large number of samples with normal sentence length. Heuristic is performed 3 times.
\textbf{Eight Metric (8M):} CASF with only the preliminary sampling phase which normalizes the score obtained by the 8 automatic metrics used in CASF and calculates the average score.
\textbf{Single Metric (SM):} CASF with only the preliminary sampling phase which uses the automatic metric used in the preliminary sampling phases of CASF.
\textbf{Online Sampling (OL):} CASF without Constrained Controller.
We compare methods with 50\% sampling rate. 
Results for other sampling ratios are in Different Sampling Ratio section of Appendix.
In addition, the number of phases and the sampling ratio of each phase are 5 and 10\%. The determination of these parameters is shown in the Phases and Associated Sampling Ratios section.
We also treat the sample size as an independent variable and results are shown in the Appendix.

\section{Results and Analysis}
\begin{table}[t]
\centering
\small
\begin{tabular}{@{}cllllll@{}}
\toprule
Method & \multicolumn{1}{c}{SUM} & \multicolumn{1}{c}{MT} & \multicolumn{1}{c}{DGen} & \multicolumn{1}{c}{SGen} & \multicolumn{1}{c}{MMGen} & \multicolumn{1}{c}{Overall} \\ \hline
R & 0.76 & 0.87 & 0.78 & 0.67 & 1.00 & 0.76 \\
H & 0.80 & 0.67 & 0.78 & 0.67 & 1.00 & 0.78 \\
8M & 0.83 & 0.80 & 0.83 & 1.00 & 1.00 & 0.84 \\
SM & 0.90 & 1.00 & 0.83 & 1.00 & 1.00 & 0.91 \\
OL & 0.69 & 0.80 & 1.00 & 1.00 & 1.00 & 0.77 \\
CASF & 0.93 & 0.80 & 1.00 & 1.00 & 1.00 & \textbf{0.93} \\ \bottomrule
\end{tabular}
\caption{Top-ranked accuracy on 5 NLG tasks. ‘Overall’ shows the average result on all human metrics from all tasks. 
}
\label{tab:res-top1}
\end{table}

\subsection{Comparison Results}

\subsubsection{Full Inter-System Ranking Accuracy}
According to results on validation set (Automatic Metrics for Preliminary Sampling Phase section of Appendix), We select MOVER-SCORE \cite{zhao2019moverscore} for calculating sample quality in the preliminary sampling phase. 
Inter-system ranking accuracy of methods on 16 datasets across 5 NLG tasks are shown in Table \ref{tab:result}. 
The results show Random have large fluctuations. For example, in the newstest2020 cn-en dataset of MT task, different times of random sampling result in different inter-system correlation. This shows the risky of widely using Random in evaluation.
CASF ranked first on 79.55\% of human metrics and ranked first or ranked second on 90.91\% of metrics. This shows CASF can better select representative samples to get a more accurate ranking. Results of the remaining human metrics, although not ranking first, are still acceptable and close to the best results. These acceptable results appear as we measure the quality of each sample in the dataset. However, human evaluation in different aspects is conducted in the same dataset. The overall scores can represent the overall evaluation results. 
We use Wilcoxon signed ranks \shortcite{demvsar2006statistical} to test the results of Random and Heuristic (both iterated 10000 times) with CASF in 44 human metrics. Results show CASF is statistically outperforming Random, Heuristic and other methods with $p = 0.00010$, $p = 0.00009$ and $p < 0.05$. 

\subsubsection{Top-Ranked System Accuracy}
One of the important goals of evaluation is to select the top-ranked system. Accurately selecting the best system with limited manpower can help the NLG field to keep good systems and eliminate poor ones. Thus, we explore the ability of CASF to identify the top-ranked system. As shown in Table \ref{tab:res-top1}, CASF achieves 93.18\% top-ranked system recognition accuracy in 44 human evaluation metrics involving 137 NLG systems.
For typical NLG tasks like DGen, SGen and MMGen, CASF achieves 100\% identification accuracy.
Experimental results also showed CASF was statistically outperforming the popular Random and Heuristic at the $p < 0.05$ level.

\begin{table*}[]
\centering
\small
\begin{tabular}{ccccc|ccccc|ccccc|ccccc}
\toprule
\textbf{M} & \textbf{\#P} & \textbf{P-R} & \textbf{B-R} & \textbf{Tau} & \textbf{M} & \textbf{\#P} & \textbf{P-R} & \textbf{B-R} & \textbf{Tau} & \textbf{M} & \textbf{\#P} & \textbf{P-R} & \textbf{B-R} & \textbf{Tau} & \textbf{M} & \textbf{\#P} & \textbf{P-R} & \textbf{B-R} & \textbf{Tau} \\ \hline
\multirow{9}{*}{A} & 2 & 0.25 & 0.25 & 0.75 & \multirow{9}{*}{F} & 2 & 0.10 & 0.40 & 0.73 & \multirow{9}{*}{F} & 2 & 0.05 & 0.45 & 0.74 & \multirow{9}{*}{F} & 2 & 0.15 & 0.35 & 0.73 \\
 & 3 & 0.17 & 0.17 & 0.76 &  & 3 & 0.10 & 0.20 & 0.75 &  & 3 & 0.05 & 0.23 & 0.74 &  & 3 & 0.15 & 0.18 & 0.77 \\
 & 4 & 0.13 & 0.13 & 0.76 &  & 4 & 0.10 & 0.13 & 0.80 &  & 4 & 0.05 & 0.15 & 0.77 &  & 4 & 0.15 & 0.12 & 0.76 \\
 & \textbf{5} & 0.10 & 0.10 & \textbf{0.83} &  & \textbf{5} & 0.10 & 0.10 & \textbf{0.83} &  & \textbf{5} & 0.05 & 0.11 & \textbf{0.77} &  & 5 & 0.15 & 0.09 & 0.76 \\
 & 6 & 0.08 & 0.08 & 0.72 &  & 6 & 0.10 & 0.08 & 0.75 &  & 6 & 0.05 & 0.09 & 0.73 &  & 6 & 0.15 & 0.07 & 0.71 \\
 & 7 & 0.07 & 0.07 & 0.72 &  & 7 & 0.10 & 0.07 & 0.69 &  & 7 & 0.05 & 0.08 & 0.69 &  & 7 & 0.15 & 0.06 & 0.73 \\
 & 8 & 0.06 & 0.06 & 0.70 &  & 8 & 0.10 & 0.06 & 0.73 &  & 8 & 0.05 & 0.06 & 0.72 &  & \textbf{8} & 0.15 & 0.05 & \textbf{0.79} \\
 & 9 & 0.06 & 0.06 & 0.73 &  & 9 & 0.10 & 0.05 & 0.72 &  & 9 & 0.05 & 0.06 & 0.72 &  & 9 & 0.15 & 0.04 & 0.75 \\
 & 10 & 0.05 & 0.05 & 0.75 &  & 10 & 0.10 & 0.04 & 0.73 &  & 10 & 0.05 & 0.05 & 0.75 &  & 10 & 0.15 & 0.04 & 0.70 \\ \bottomrule
\end{tabular}
\caption{Experimental results on 44 human metrics with different mode (M) (Average (A) and Preliminary-Fixed (F)), number of phases (\#P), preliminary sample ratio (P-R) and batch sampling ratio (B-R) of each phase for the proposed CASF.}
\label{tab:paraSelect}
\end{table*} 

\subsubsection{Case Study}
\begin{figure}[t]
    \centering
    \includegraphics[width=0.45\textwidth,height=0.36\textwidth]{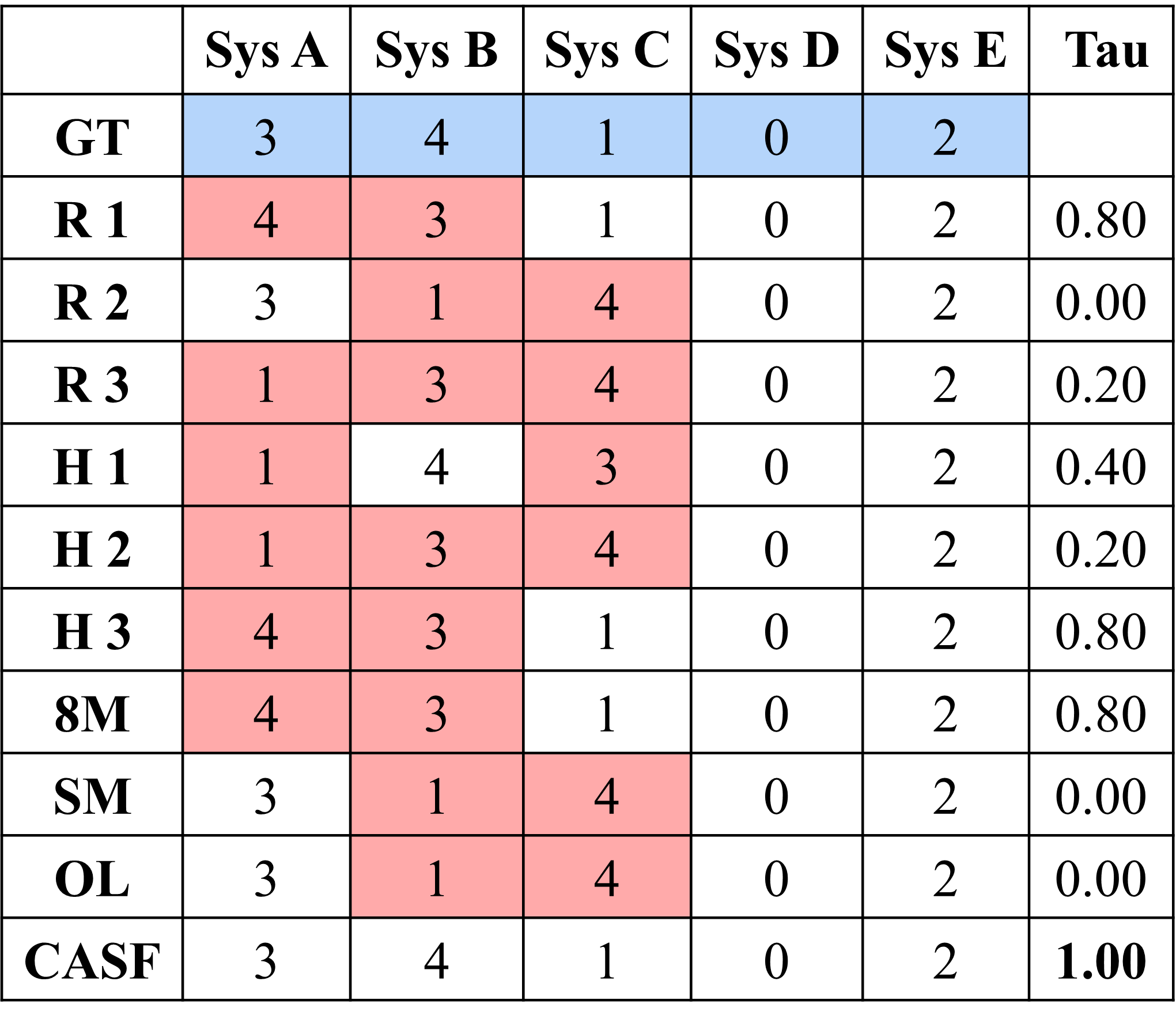}
    \caption{Inter-system ranking of human evaluation aspect `accuracy' of OpenAI 1. ``GT" is the inter-system ranking on the entire dataset. Sampling rate is 50\%. ``Sys" represents system. Rankings in red indicate incorrect rankings.}
    \label{fig:caseStudy}
\end{figure} 
Taking the human aspect accuracy in the OpenAI 1 \cite{stiennon2020learning,volske2017tl} dataset as an example, CASF obtains an accurate inter-system ranking as shown in Figure \ref{fig:caseStudy}. 
The 3 times of random sampling obtained different inter-system rankings, and the ranking of the first system fluctuated between the first and fourth, with great volatility. This confirms the problem we raised about the risk of random sampling, making  evaluation unreliable.
CASF selects the same subset in multiple times, and the variance of the inter-ranking accuracy obtained by multiple sampling times is 0 (Learner Selection section of Appendix).
Since CASF selects representative samples, it obtains more accurate inter-system rankings, making evaluation more reliable.

\subsection{Automatic Metric for Preliminary Phase}\label{sec:auto-metric}
\begin{table}[b]
\centering
\small
\begin{tabular}{@{}ccccccc@{}}
\toprule
\textbf{Metric} & SUM & MT & DGen & SGen & \multicolumn{1}{c}{MMGen} & Avg \\ \hline
BERT-S & 0.74 & 0.58 & 0.67 & 1.00 & 1.00 & 0.73 \\
MOVER-S & 0.84 & 0.58 & 0.89 & 1.00 & 1.00 & \textbf{0.83} \\
ROUGE-1 & 0.73 & 0.57 & 0.67 & 0.30 & 1.00 & 0.70 \\
ROUGE-2 & 0.73 & 0.55 & 0.56 & 1.00 & 0.80 & 0.70 \\
ROUGE-L & 0.72 & 0.52 & 0.89 & 1.00 & 1.00 & 0.75 \\
BART-S & 0.60 & 0.44 & 0.89 & 0.90 & 0.80 & 0.64 \\
BLEU & 0.72 & 0.37 & 0.56 & 1.00 & 0.80 & 0.67 \\
METEOR & 0.78 & 0.54 & 0.89 & 1.00 & 1.00 & {\underline{0.79}}\\ \bottomrule
\end{tabular}
\caption{Results of CASF pre-ranking on different automatic metrics. ``-S" indicates ``-Score".
``Avg" represents the average result on all human metrics from all tasks.
}
\label{tab:res-metrics}
\end{table}
We choose automatic metrics commonly used in NLG as our automatic metrics set, including BERT-SCORE \shortcite{zhang2019bertscore}, MOVER-SCORE \shortcite{zhao2019moverscore}, ROUGE-1 \shortcite{lin2004rouge}, 	ROUGE-2, ROUGE-L, BART-SCORE \shortcite{yuan2021bartscore}, BLEU \shortcite{papineni2002bleu} and METEOR \shortcite{banerjee2005meteor}. 
We apply each metric to calculate sample quality in the preliminary sampling phase of CASF in Table \ref{tab:res-metrics}.
Results show sample quality calculated on MOVER-SCORE get a more correct  ranking. This shows the ability to calculate sample quality of contextual-embedding-based metric MOVER-SCORE. 
Traditional metric METEOR ranks second.
Full results are in Appendix.

\subsection{Phases and Associated Sampling Ratios}\label{sec:ParaSelect}
We conduct experiments to explore the influence of the number of phases and the sampling ratio of each phase for CASF.
Results at the sampling rate of 50\% on 16 datasets are shown in Table \ref{tab:paraSelect}. 
In average mode, all phases are sampled in equal proportions. In the preliminary-fixed mode, we fix the preliminary sampling ratio, and the batch sampling ratio is divided equally according to the number of iteration phases and the total sampling ratio. 
Results show that performance is better when the number of iteration phases is 5 in most cases.
It is simple and effective to sample each phase according to the total sampling rate and the number of phases.

\subsection{Significant Information Retention Accuracy}
Previous work \shortcite{mohankumar2022active} focused on identifying top-ranked systems, and we further explored giving more accurate overall inter-system rankings
and tested the significant information retention accuracy on sample subsets, that is, to test whether the subset can preserve the significance of ranking among systems. Results showed CASF outperforms Random and Heuristic. Details are in the Appendix.


\section{Related Work}
Previous works \shortcite{bojar2014findings,bojar-etal-2015-findings,sakaguchi2014efficient,sakaguchi2016reassessing} adopt TrueSkill \shortcite{herbrich2006trueskill} to rank NLG methods with pairwise human evaluation.  
\citet{sakaguchi2018efficient} introduce a method for system quality estimation from pairwise annotation by human judgment. 
Hashimoto et al. \shortcite{hashimoto2019unifying} propose an evaluation mechanism to calculate a model's sampling probabilities.
\citet{chaganty2018price} utilize control variates to obtain an unbiased estimator with lower cost than only using human evaluation.
\citet{mendoncca2021online} adopt online learning to find the best systems for machine translation. 
Wei et al. \shortcite{wei2022searching} study the power on pairwise direct assessment comparisons.
A recent work \shortcite{mohankumar2022active} introduces Active Evaluation to identify the top-ranked system with less pairwise human annotations. 
There is still a vacancy in the research to derive a complete inter-system ranking based on the results of direct human scoring for general NLG tasks.
\citet{yates1948systematic} proposed Systematic Sampling.
ILDAE \shortcite{varshney2022ildae} calculates the difficulty score of the sample and uses a simple sampling method for Natural Language Inference. However, ILDAE is not suitable for NLG since there is no direct confidence value in NLG methods. To the best of our knowledge, this paper is the first work to extensively study the sampling method for direct scoring to get the whole inter-system ranking in NLG human evaluation.
\section{Conclusion}
In this paper, we focused on giving a more correct inter-system ranking for reliable human evaluation with limited time and cost. We propose CASF and show the overall inter-system Kendall correlation improved by 41\% to 0.83 compared to the widely used random sampling in 44 human evaluation metrics across 16 datasets in 5 NLG tasks. CASF ranked first or ranked second among all comparison methods on up to 90.91\% of the human metrics. We release a tool and we strongly recommend using CASF for reliable human evaluation 
to get a more reliable inter-system ranking.

\section{Acknowledgements}
This work was supported by National Key R\&D Program of China (2021YFF0901502), National Science Foundation of China (No. 62161160339), State Key Laboratory of Media Convergence Production Technology and Systems and Key Laboratory of Science, Technology and Standard in Press Industry (Key Laboratory of Intelligent Press Media Technology). We appreciate the anonymous reviewers for their helpful comments. Xiaojun Wan is the corresponding author.



\bibliography{main}

\begin{thebibliography}{60}
\providecommand{\natexlab}[1]{#1}

\bibitem[{Abdi et~al.(2007)}]{abdi2007method}
Abdi, H.; et~al. 2007.
\newblock The method of least squares.
\newblock \emph{Encyclopedia of measurement and statistics}, 1: 530--532.

\bibitem[{Banerjee and Lavie(2005)}]{banerjee2005meteor}
Banerjee, S.; and Lavie, A. 2005.
\newblock METEOR: An automatic metric for MT evaluation with improved correlation with human judgments.
\newblock In \emph{Proceedings of the acl workshop on intrinsic and extrinsic evaluation measures for machine translation and/or summarization}, 65--72.

\bibitem[{Begg and Mazumdar(1994)}]{begg1994operating}
Begg, C.~B.; and Mazumdar, M. 1994.
\newblock Operating characteristics of a rank correlation test for publication bias.
\newblock \emph{Biometrics}, 1088--1101.

\bibitem[{Bethard(2022)}]{bethard2022we}
Bethard, S. 2022.
\newblock We need to talk about random seeds.
\newblock \emph{arXiv preprint arXiv:2210.13393}.

\bibitem[{Bhandari et~al.(2020)Bhandari, Gour, Ashfaq, Liu, and Neubig}]{bhandari2020re}
Bhandari, M.; Gour, P.; Ashfaq, A.; Liu, P.; and Neubig, G. 2020.
\newblock Re-evaluating evaluation in text summarization.
\newblock \emph{arXiv preprint arXiv:2010.07100}.

\bibitem[{Bhatnagar, Ganesh, and Kann(2022)}]{bhatnagar2022chia}
Bhatnagar, R.; Ganesh, A.; and Kann, K. 2022.
\newblock CHIA: CHoosing Instances to Annotate for Machine Translation.
\newblock In \emph{Findings of the Association for Computational Linguistics: EMNLP 2022}, 7299--7315.

\bibitem[{Bojar et~al.(2014)Bojar, Buck, Federmann, Haddow, Koehn, Leveling, Monz, Pecina, Post, Saint-Amand et~al.}]{bojar2014findings}
Bojar, O.; Buck, C.; Federmann, C.; Haddow, B.; Koehn, P.; Leveling, J.; Monz, C.; Pecina, P.; Post, M.; Saint-Amand, H.; et~al. 2014.
\newblock Findings of the 2014 workshop on statistical machine translation.
\newblock In \emph{Proceedings of the ninth workshop on statistical machine translation}, 12--58.

\bibitem[{Bojar et~al.(2015)Bojar, Chatterjee, Federmann, Haddow, Huck, Hokamp, Koehn, Logacheva, Monz, Negri, Post, Scarton, Specia, and Turchi}]{bojar-etal-2015-findings}
Bojar, O.; Chatterjee, R.; Federmann, C.; Haddow, B.; Huck, M.; Hokamp, C.; Koehn, P.; Logacheva, V.; Monz, C.; Negri, M.; Post, M.; Scarton, C.; Specia, L.; and Turchi, M. 2015.
\newblock Findings of the 2015 Workshop on Statistical Machine Translation.
\newblock In \emph{Proceedings of the Tenth Workshop on Statistical Machine Translation}, 1--46. Lisbon, Portugal: Association for Computational Linguistics.

\bibitem[{Breiman(1996)}]{breiman1996bagging}
Breiman, L. 1996.
\newblock Bagging predictors.
\newblock \emph{Machine learning}, 24(2): 123--140.

\bibitem[{Breiman(2001)}]{breiman2001random}
Breiman, L. 2001.
\newblock Random forests.
\newblock \emph{Machine learning}, 45(1): 5--32.

\bibitem[{Card et~al.(2020)Card, Henderson, Khandelwal, Jia, Mahowald, and Jurafsky}]{card2020little}
Card, D.; Henderson, P.; Khandelwal, U.; Jia, R.; Mahowald, K.; and Jurafsky, D. 2020.
\newblock With Little Power Comes Great Responsibility.
\newblock In \emph{Proceedings of the 2020 Conference on Empirical Methods in Natural Language Processing (EMNLP)}, 9263--9274.

\bibitem[{Celikyilmaz, Clark, and Gao(2020)}]{celikyilmaz2020evaluation}
Celikyilmaz, A.; Clark, E.; and Gao, J. 2020.
\newblock Evaluation of text generation: A survey.
\newblock \emph{arXiv preprint arXiv:2006.14799}.

\bibitem[{Chaganty, Mussman, and Liang(2018)}]{chaganty2018price}
Chaganty, A.~T.; Mussman, S.; and Liang, P. 2018.
\newblock The price of debiasing automatic metrics in natural language evaluation.
\newblock \emph{arXiv preprint arXiv:1807.02202}.

\bibitem[{Cover and Hart(1967)}]{cover1967nearest}
Cover, T.; and Hart, P. 1967.
\newblock Nearest neighbor pattern classification.
\newblock \emph{IEEE transactions on information theory}, 13(1): 21--27.

\bibitem[{Dem{\v{s}}ar(2006)}]{demvsar2006statistical}
Dem{\v{s}}ar, J. 2006.
\newblock Statistical comparisons of classifiers over multiple data sets.
\newblock \emph{The Journal of Machine learning research}, 7: 1--30.

\bibitem[{Du{\v{s}}ek, Novikova, and Rieser(2018)}]{dusek-etal-2018-findings}
Du{\v{s}}ek, O.; Novikova, J.; and Rieser, V. 2018.
\newblock Findings of the {E}2{E} {NLG} Challenge.
\newblock In \emph{Proceedings of the 11th International Conference on Natural Language Generation}, 322--328. Tilburg University, The Netherlands: Association for Computational Linguistics.

\bibitem[{Fabbri et~al.(2021)Fabbri, Kry{\'s}ci{\'n}ski, McCann, Xiong, Socher, and Radev}]{fabbri2021summeval}
Fabbri, A.~R.; Kry{\'s}ci{\'n}ski, W.; McCann, B.; Xiong, C.; Socher, R.; and Radev, D. 2021.
\newblock Summeval: Re-evaluating summarization evaluation.
\newblock \emph{Transactions of the Association for Computational Linguistics}, 9: 391--409.

\bibitem[{Freitag et~al.(2021)Freitag, Foster, Grangier, Ratnakar, Tan, and Macherey}]{freitag2021experts}
Freitag, M.; Foster, G.; Grangier, D.; Ratnakar, V.; Tan, Q.; and Macherey, W. 2021.
\newblock Experts, Errors, and Context: A Large-Scale Study of Human Evaluation for Machine Translation.
\newblock arXiv:2104.14478.

\bibitem[{Freund, Schapire et~al.(1996)}]{freund1996experiments}
Freund, Y.; Schapire, R.~E.; et~al. 1996.
\newblock Experiments with a new boosting algorithm.
\newblock In \emph{icml}, volume~96, 148--156. Citeseer.

\bibitem[{Friedman(2001)}]{friedman2001greedy}
Friedman, J.~H. 2001.
\newblock Greedy function approximation: a gradient boosting machine.
\newblock \emph{Annals of statistics}, 1189--1232.

\bibitem[{Gao and Wan(2022)}]{gao-wan-2022-dialsummeval}
Gao, M.; and Wan, X. 2022.
\newblock {D}ial{S}umm{E}val: Revisiting Summarization Evaluation for Dialogues.
\newblock In \emph{Proceedings of the 2022 Conference of the North American Chapter of the Association for Computational Linguistics: Human Language Technologies}, 5693--5709. Seattle, United States: Association for Computational Linguistics.

\bibitem[{Gatt and Krahmer(2018)}]{gatt2018survey}
Gatt, A.; and Krahmer, E. 2018.
\newblock Survey of the state of the art in natural language generation: Core tasks, applications and evaluation.
\newblock \emph{Journal of Artificial Intelligence Research}, 61: 65--170.

\bibitem[{Geurts, Ernst, and Wehenkel(2006)}]{geurts2006extremely}
Geurts, P.; Ernst, D.; and Wehenkel, L. 2006.
\newblock Extremely randomized trees.
\newblock \emph{Machine learning}, 63(1): 3--42.

\bibitem[{Gkatzia and Mahamood(2015)}]{gkatzia2015snapshot}
Gkatzia, D.; and Mahamood, S. 2015.
\newblock A snapshot of NLG evaluation practices 2005-2014.
\newblock In \emph{Proceedings of the 15th European Workshop on Natural Language Generation (ENLG)}, 57--60.

\bibitem[{Grusky, Naaman, and Artzi(2018)}]{grusky2018newsroom}
Grusky, M.; Naaman, M.; and Artzi, Y. 2018.
\newblock Newsroom: A dataset of 1.3 million summaries with diverse extractive strategies.
\newblock \emph{arXiv preprint arXiv:1804.11283}.

\bibitem[{Guan et~al.(2021)Guan, Zhang, Feng, Liu, Ding, Mao, Fan, and Huang}]{guan2021openmeva}
Guan, J.; Zhang, Z.; Feng, Z.; Liu, Z.; Ding, W.; Mao, X.; Fan, C.; and Huang, M. 2021.
\newblock OpenMEVA: A Benchmark for Evaluating Open-ended Story Generation Metrics.
\newblock arXiv:2105.08920.

\bibitem[{Hashimoto, Zhang, and Liang(2019)}]{hashimoto2019unifying}
Hashimoto, T.~B.; Zhang, H.; and Liang, P. 2019.
\newblock Unifying human and statistical evaluation for natural language generation.
\newblock \emph{arXiv preprint arXiv:1904.02792}.

\bibitem[{Hearst et~al.(1998)Hearst, Dumais, Osuna, Platt, and Scholkopf}]{hearst1998support}
Hearst, M.~A.; Dumais, S.~T.; Osuna, E.; Platt, J.; and Scholkopf, B. 1998.
\newblock Support vector machines.
\newblock \emph{IEEE Intelligent Systems and their applications}, 13(4): 18--28.

\bibitem[{Herbrich, Minka, and Graepel(2006)}]{herbrich2006trueskill}
Herbrich, R.; Minka, T.; and Graepel, T. 2006.
\newblock TrueSkill™: a Bayesian skill rating system.
\newblock \emph{Advances in neural information processing systems}, 19.

\bibitem[{Howcroft et~al.(2020)Howcroft, Belz, Clinciu, Gkatzia, Hasan, Mahamood, Mille, Van~Miltenburg, Santhanam, and Rieser}]{howcroft2020twenty}
Howcroft, D.~M.; Belz, A.; Clinciu, M.-A.; Gkatzia, D.; Hasan, S.~A.; Mahamood, S.; Mille, S.; Van~Miltenburg, E.; Santhanam, S.; and Rieser, V. 2020.
\newblock Twenty years of confusion in human evaluation: NLG needs evaluation sheets and standardised definitions.
\newblock In \emph{Proceedings of the 13th International Conference on Natural Language Generation}, 169--182.

\bibitem[{Hu et~al.(2019)Hu, Rudinger, Post, and Van~Durme}]{hu2019parabank}
Hu, J.~E.; Rudinger, R.; Post, M.; and Van~Durme, B. 2019.
\newblock Parabank: Monolingual bitext generation and sentential paraphrasing via lexically-constrained neural machine translation.
\newblock In \emph{Proceedings of the AAAI Conference on Artificial Intelligence}, volume~33, 6521--6528.

\bibitem[{Kasai et~al.(2022)Kasai, Sakaguchi, Dunagan, Morrison, Bras, Choi, and Smith}]{kasai2022thumb}
Kasai, J.; Sakaguchi, K.; Dunagan, L.; Morrison, J.; Bras, R.~L.; Choi, Y.; and Smith, N.~A. 2022.
\newblock Transparent Human Evaluation for Image Captioning.
\newblock In \emph{Proc.\ of NAACL}.

\bibitem[{Kendall(1938)}]{kendall1938new}
Kendall, M.~G. 1938.
\newblock A new measure of rank correlation.
\newblock \emph{Biometrika}, 30(1/2): 81--93.

\bibitem[{Kendall(1945)}]{kendall1945treatment}
Kendall, M.~G. 1945.
\newblock The treatment of ties in ranking problems.
\newblock \emph{Biometrika}, 33(3): 239--251.

\bibitem[{Kondrak(2005)}]{kondrak2005n}
Kondrak, G. 2005.
\newblock N-gram similarity and distance.
\newblock In \emph{International symposium on string processing and information retrieval}, 115--126. Springer.

\bibitem[{Lin(2004)}]{lin2004rouge}
Lin, C.-Y. 2004.
\newblock Rouge: A package for automatic evaluation of summaries.
\newblock In \emph{Text summarization branches out}, 74--81.

\bibitem[{Mehri and Eskenazi(2020)}]{mehri2020usr}
Mehri, S.; and Eskenazi, M. 2020.
\newblock USR: An unsupervised and reference free evaluation metric for dialog generation.
\newblock \emph{arXiv preprint arXiv:2005.00456}.

\bibitem[{Mendon{\c{c}}a et~al.(2021)Mendon{\c{c}}a, Rei, Coheur, Sardinha, and Santos}]{mendoncca2021online}
Mendon{\c{c}}a, V.; Rei, R.; Coheur, L.; Sardinha, A.; and Santos, A.~L. 2021.
\newblock Online learning meets machine translation evaluation: Finding the best systems with the least human effort.
\newblock \emph{arXiv preprint arXiv:2105.13385}.

\bibitem[{Mohankumar and Khapra(2022)}]{mohankumar2022active}
Mohankumar, A.~K.; and Khapra, M.~M. 2022.
\newblock Active Evaluation: Efficient NLG Evaluation with Few Pairwise Comparisons.
\newblock In \emph{Proceedings of the 60th Annual Meeting of the Association for Computational Linguistics (Volume 1: Long Papers)}, 8761--8781.

\bibitem[{Novikova et~al.(2017)Novikova, Du{\v{s}}ek, Curry, and Rieser}]{novikova2017we}
Novikova, J.; Du{\v{s}}ek, O.; Curry, A.~C.; and Rieser, V. 2017.
\newblock Why we need new evaluation metrics for NLG.
\newblock \emph{arXiv preprint arXiv:1707.06875}.

\bibitem[{Papineni et~al.(2002)Papineni, Roukos, Ward, and Zhu}]{papineni2002bleu}
Papineni, K.; Roukos, S.; Ward, T.; and Zhu, W.-J. 2002.
\newblock Bleu: a method for automatic evaluation of machine translation.
\newblock In \emph{Proceedings of the 40th annual meeting of the Association for Computational Linguistics}, 311--318.

\bibitem[{Peyrard(2019)}]{peyrard2019simple}
Peyrard, M. 2019.
\newblock A Simple Theoretical Model of Importance for Summarization.
\newblock In \emph{Proceedings of the 57th Annual Meeting of the Association for Computational Linguistics}, 1059--1073.

\bibitem[{Quinlan(1986)}]{quinlan1986induction}
Quinlan, J.~R. 1986.
\newblock Induction of decision trees.
\newblock \emph{Machine learning}, 1(1): 81--106.

\bibitem[{Reiter and Belz(2009)}]{reiter2009investigation}
Reiter, E.; and Belz, A. 2009.
\newblock An investigation into the validity of some metrics for automatically evaluating natural language generation systems.
\newblock \emph{Computational Linguistics}, 35(4): 529--558.

\bibitem[{Sakaguchi et~al.(2016)Sakaguchi, Napoles, Post, and Tetreault}]{sakaguchi2016reassessing}
Sakaguchi, K.; Napoles, C.; Post, M.; and Tetreault, J. 2016.
\newblock Reassessing the goals of grammatical error correction: Fluency instead of grammaticality.
\newblock \emph{Transactions of the Association for Computational Linguistics}, 4: 169--182.

\bibitem[{Sakaguchi, Post, and Van~Durme(2014)}]{sakaguchi2014efficient}
Sakaguchi, K.; Post, M.; and Van~Durme, B. 2014.
\newblock Efficient elicitation of annotations for human evaluation of machine translation.
\newblock In \emph{Proceedings of the Ninth Workshop on Statistical Machine Translation}, 1--11.

\bibitem[{Sakaguchi and Van~Durme(2018)}]{sakaguchi2018efficient}
Sakaguchi, K.; and Van~Durme, B. 2018.
\newblock Efficient online scalar annotation with bounded support.
\newblock \emph{arXiv preprint arXiv:1806.01170}.

\bibitem[{Seber and Lee(2012)}]{seber2012linear}
Seber, G.~A.; and Lee, A.~J. 2012.
\newblock \emph{Linear regression analysis}.
\newblock John Wiley \& Sons.

\bibitem[{Shi et~al.(2022)Shi, Yang, Xu, Yuan, Li, Hu, and Zha}]{DBLP:conf/cvpr/ShiYXYLHZ22}
Shi, Y.; Yang, X.; Xu, H.; Yuan, C.; Li, B.; Hu, W.; and Zha, Z. 2022.
\newblock EMScore: Evaluating Video Captioning via Coarse-Grained and Fine-Grained Embedding Matching.
\newblock In \emph{{IEEE/CVF} Conference on Computer Vision and Pattern Recognition, {CVPR} 2022, New Orleans, LA, USA, June 18-24, 2022}.

\bibitem[{Stiennon et~al.(2020)Stiennon, Ouyang, Wu, Ziegler, Lowe, Voss, Radford, Amodei, and Christiano}]{stiennon2020learning}
Stiennon, N.; Ouyang, L.; Wu, J.; Ziegler, D.; Lowe, R.; Voss, C.; Radford, A.; Amodei, D.; and Christiano, P.~F. 2020.
\newblock Learning to summarize with human feedback.
\newblock \emph{Advances in Neural Information Processing Systems}, 33: 3008--3021.

\bibitem[{Varshney, Mishra, and Baral(2022)}]{varshney2022ildae}
Varshney, N.; Mishra, S.; and Baral, C. 2022.
\newblock ILDAE: Instance-Level Difficulty Analysis of Evaluation Data.
\newblock \emph{arXiv preprint arXiv:2203.03073}.

\bibitem[{V{\"o}lske et~al.(2017)V{\"o}lske, Potthast, Syed, and Stein}]{volske2017tl}
V{\"o}lske, M.; Potthast, M.; Syed, S.; and Stein, B. 2017.
\newblock Tl; dr: Mining reddit to learn automatic summarization.
\newblock In \emph{Proceedings of the Workshop on New Frontiers in Summarization}, 59--63.

\bibitem[{Wan and Xiao(2008)}]{wan2008collabrank}
Wan, X.; and Xiao, J. 2008.
\newblock CollabRank: towards a collaborative approach to single-document keyphrase extraction.
\newblock In \emph{Proceedings of the 22nd International Conference on Computational Linguistics (Coling 2008)}, 969--976.

\bibitem[{Wan and Yang(2007)}]{wan2007collabsum}
Wan, X.; and Yang, J. 2007.
\newblock CollabSum: exploiting multiple document clustering for collaborative single document summarizations.
\newblock In \emph{Proceedings of the 30th annual international ACM SIGIR conference on Research and development in information retrieval}, 143--150.

\bibitem[{Wei, Kocmi, and Federmann(2022)}]{wei2022searching}
Wei, J. T.-Z.; Kocmi, T.; and Federmann, C. 2022.
\newblock Searching for a higher power in the human evaluation of MT.
\newblock \emph{arXiv preprint arXiv:2210.11612}.

\bibitem[{Yates(1948)}]{yates1948systematic}
Yates, F. 1948.
\newblock Systematic sampling.
\newblock \emph{Philosophical Transactions of the Royal Society of London. Series A, Mathematical and Physical Sciences}, 241(834): 345--377.

\bibitem[{Yuan, Neubig, and Liu(2021)}]{yuan2021bartscore}
Yuan, W.; Neubig, G.; and Liu, P. 2021.
\newblock Bartscore: Evaluating generated text as text generation.
\newblock \emph{Advances in Neural Information Processing Systems}, 34: 27263--27277.

\bibitem[{Zhang et~al.(2019)Zhang, Kishore, Wu, Weinberger, and Artzi}]{zhang2019bertscore}
Zhang, T.; Kishore, V.; Wu, F.; Weinberger, K.~Q.; and Artzi, Y. 2019.
\newblock Bertscore: Evaluating text generation with bert.
\newblock \emph{arXiv preprint arXiv:1904.09675}.

\bibitem[{Zhao et~al.(2019)Zhao, Peyrard, Liu, Gao, Meyer, and Eger}]{zhao2019moverscore}
Zhao, W.; Peyrard, M.; Liu, F.; Gao, Y.; Meyer, C.~M.; and Eger, S. 2019.
\newblock MoverScore: Text generation evaluating with contextualized embeddings and earth mover distance.
\newblock \emph{arXiv preprint arXiv:1909.02622}.

\bibitem[{Zhou et~al.(2022)Zhou, Blodgett, Trischler, Daum{\'e}~III, Suleman, and Olteanu}]{zhou2022deconstructing}
Zhou, K.; Blodgett, S.~L.; Trischler, A.; Daum{\'e}~III, H.; Suleman, K.; and Olteanu, A. 2022.
\newblock Deconstructing NLG Evaluation: Evaluation Practices, Assumptions, and Their Implications.
\newblock \emph{arXiv preprint arXiv:2205.06828}.

\end{thebibliography}

\appendix
\clearpage
\section{Appendix}
\label{sec:appendix}

\subsection{Survey}
\begin{figure}[h]
    \centering
    \includegraphics[scale=0.35]{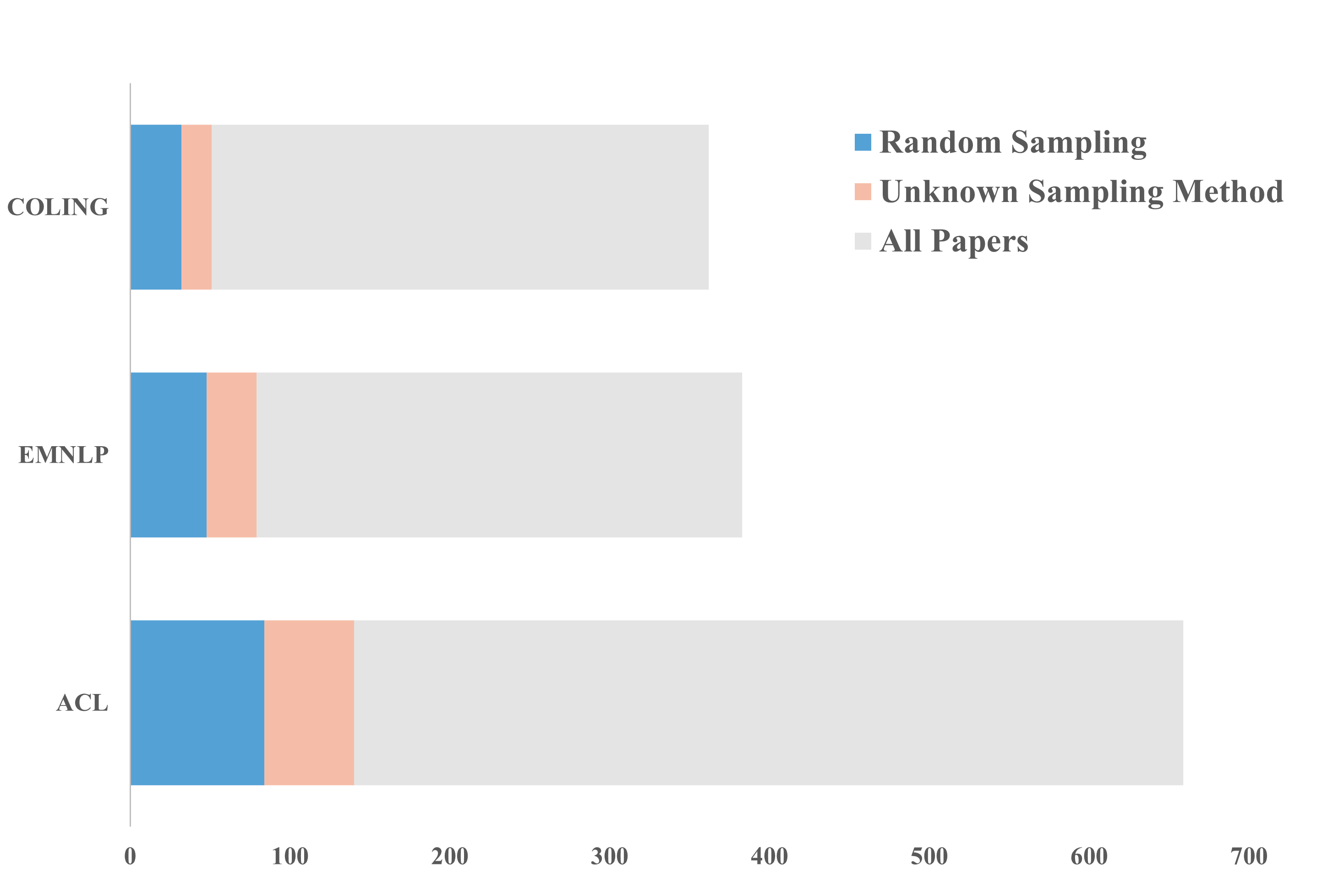}
    \caption{The number of papers that use random sampling and do not mention the sampling method they used (unknown sampling methods) for human evaluation in top NLP conference.}
    \label{fig:survey1}
\end{figure}

\begin{figure}[h]
    \centering
    \includegraphics[scale=0.4]{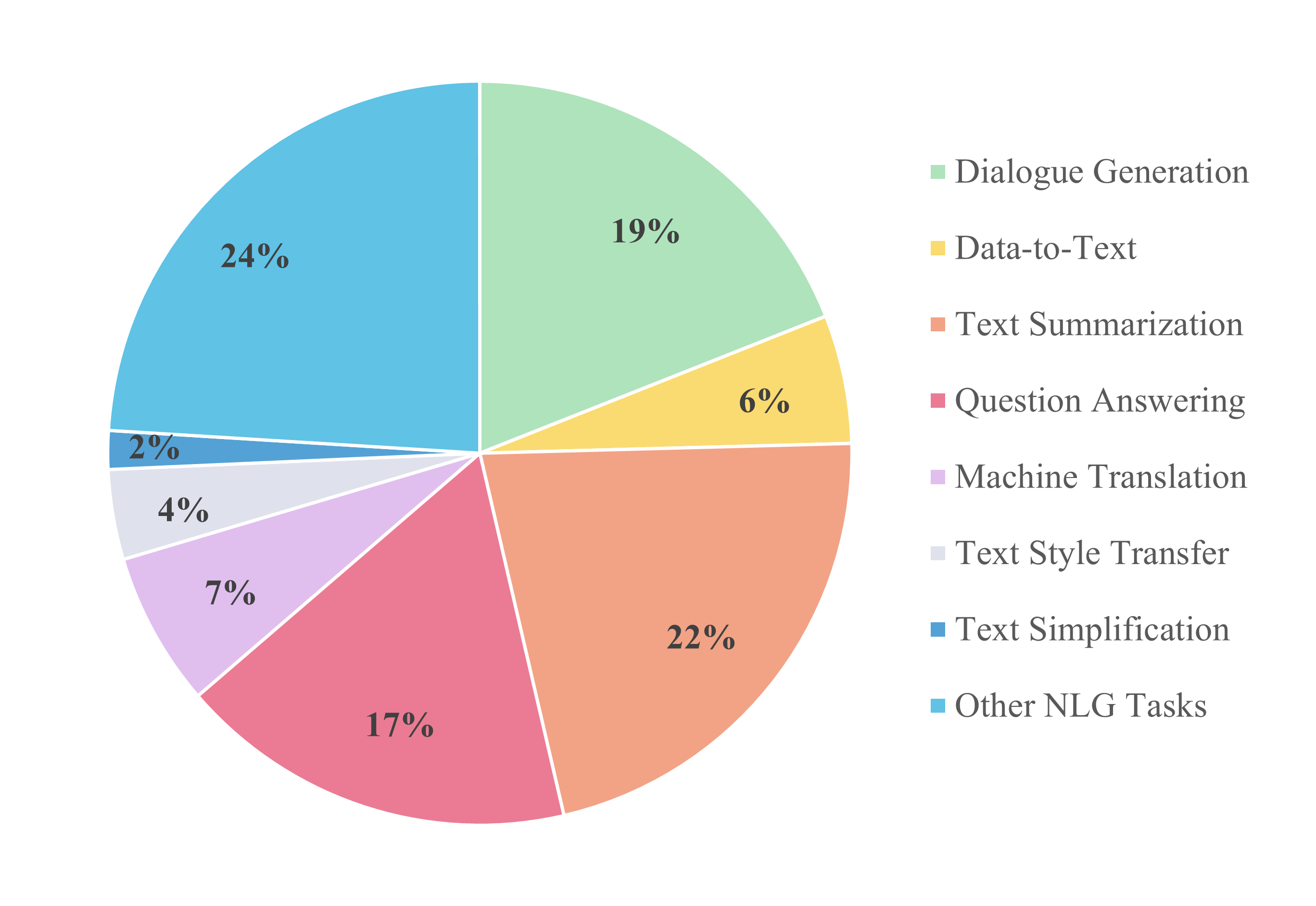}
    \caption{Proportion of NLG papers sampling a subset to conduct manual evaluation on different NLG tasks.}
    \label{fig:survey2}
\end{figure}

We investigate papers with human evaluation to better study the current manual evaluation sampling problem. First, we randomly selected 1404 papers from ACL, EMNLP and COLING in the last two years from paperswithcode.com and the lists of accepted work published by the conferences. We then browse each paper by searching with the keywords `human', `manual' and `annotate', and the content of the keywords in context is viewed. We find that 270 papers selected a subset of the test dataset for manual evaluation to save the labor and cost of manual evaluation. For these papers, we use `sample' as the keyword to search for the sampling method used for human evaluation. It is found that random sampling is the most important sampling method, accounting for 60.7\%, and the other 39.3\% do not mention the sampling method they used. The number of papers using random sampling and unknown sampling methods in each conference is shown in Figure \ref{fig:survey1}.
Of these 270 papers that take human evaluation, 179 papers are from NLG tasks, with the most being text summarization at 22\%. The proportion of these NLG papers on different tasks is shown in Figure \ref{fig:survey2}. The lists of the 1404 papers surveyed and the 270 papers that selected
a subset of the test dataset for human evaluation will be released.

According to the results of the survey, there are two major problems in the current sampling problem of human evaluation. On the one hand, using random sampling to select samples can lead to unreliable human evaluation results, because different sampling subsets are likely to lead to different inter-system ranking results. In addition, not disclosing the list of samples selected by random sampling will lead to poor reproducibility of evaluation results. 
On the other hand, we find that up to 39.3\% of the papers do not provide information on human evaluation sampling, which would lead to low reliability and low reproducibility of evaluation results.
We recommend providing sampling information including sampling methods, evaluation sample lists, etc. when conducting human evaluations in the future to standardize the human evaluation process. At the same time, we strongly recommend using our proposed Constrained Active Sampling Framework for sampling evaluation subsets in human evaluations, which will make human evaluations more reliable and allow excellent systems to be retained.

\subsection{Tasks and Datasets} \label{ap:dataset}
\subsubsection{Test Set}
In Table \ref{tab:dataset}, we report information on the NLG tasks and related datasets we used as test set, including the number of human evaluation aspects, the number of NLG systems involved in the corresponding datasets, the sample size of the datasets and specific human evaluation aspects. The order of the human evaluation metrics in the experiment in this paper follows the order of the human metrics shown in Table \ref{tab:dataset}. As Likert-scale comparisons are the most commonly reported type of evaluation \cite{card2020little}, we focus on Likert-scale datasets. For data preprocessing, we first discard samples that lack information, including the system output, human evaluation score, and reference text. We then compute the automated metrics' scores of these samples for use.

\subsubsection{Validation Set}
We select automatic metric for the preliminary phase, regressor for the learner, number of phases and the associated sampling ratios on the validation set shown in Table \ref{tab:dataset-val}. The validation set contains nine datasets and 41 NLG systems from two traditional NLG tasks namely Data to Text and Paraphrase Generation. 
Since not all NLG systems on the E2E \cite{dusek-etal-2018-findings} and ParaBank \cite{hu2019parabank} datasets have human evaluation score on the same samples, we divide the ParaBank datasets into subsets. The samples on these subsets have human scoring results on all systems. For subsets selection, we first arranged and combined all the systems into system subsets, and calculated the number of samples that meet the need of having human evaluation score on all systems in the system subset. Then, we select the system subset with more samples that meet the need. The final system IDs of the selected subset are shown in Table \ref{tab:dataset-val}.

\begin{table*}[]
\caption{Description of tasks and datasets with the number of human evaluation metrics (\# HE Metrics), number of NLG systems (\# Systems), number of samples (\# Samples) and the human evaluation aspects (HE Metrics).}
\centering
\resizebox{\textwidth}{29mm}{
\begin{tabular}{@{}clcccl@{}}
\hline 
\textbf{Tasks} & \multicolumn{1}{c}{\textbf{Datasets}} & \textbf{\# HE Metrics} & \textbf{\# Systems} & \textbf{\# Samples} & \multicolumn{1}{c}{\textbf{HE Metrics}} \\ \hline
 & SummEval \cite{fabbri2021summeval} & 4 & 16 & 100 & coherence, consistency,   fluency,relevance \\
 & REALSumm   \cite{bhandari2020re} & 1 & 24 & 100 & litepyramid-recall \\
 & NeR18   \cite{grusky2018newsroom} & 4 & 7 & 60 & coherence, fluency, informativeness, relevance \\
 & DialSummEval   \cite{gao-wan-2022-dialsummeval} & 4 & 13 & 100 & consistency,relevance,fluency,coherence \\ 
 & OpenAI-axis1   \cite{stiennon2020learning,volske2017tl} & 4 & 5 & 439 &  accuracy,coherence,coverage,overall\\ 
 & OpenAI-axis2   \cite{stiennon2020learning,volske2017tl} & 4 & 7 & 636 &  accuracy,coherence,coverage,overall\\
 & OpenAI-CNN/DM1   \cite{stiennon2020learning,volske2017tl} & 4 & 10 & 206 & accuracy,coherence,coverage,overall \\
\multirow{-8}{*}{Summarization} & OpenAI-CNN/DM3   \cite{stiennon2020learning,volske2017tl} & 4 & 3 & 206 & accuracy,coherence,coverage,overall \\ \hline
 & {newstext2020 en-de \cite{freitag2021experts}} & 2 & 7 & 1066 & MQM, pSQM \\
 & newstext2020 cn-en   \cite{freitag2021experts} & 2 & 8 & 1641 & MQM, pSQM \\
\multirow{-3}{*}{Machine Translation} & newstext2021 cn-en   \cite{freitag2021experts} & 1 & 13 & 147 & MQM \\ \hline
\begin{tabular}[c]{@{}c@{}}Dialogue\\       Generation\end{tabular} & Persona Chat \cite{mehri2020usr} & 6 & 3 & 60 & \begin{tabular}[c]{@{}l@{}}Understandable, Natural,   Maintains Context, \\Interesting, Uses Knowledge, Overall Quality\end{tabular} \\ \hline
 & MANS-ROC \cite{guan2021openmeva} & 1 & 5 & 200 & overall \\ 
\multirow{-2}{*}{Story Generation} & MANS-WP   \cite{guan2021openmeva} & 1 & 5 & 200 & overall \\ \hline
 & THUMB-MSCOCO \cite{kasai2022thumb} & 1 & 5 & 500 & overall \\ 
\multirow{-2}{*}{Multi-Modal   Generation} & VATEX-EVAL   \cite{DBLP:conf/cvpr/ShiYXYLHZ22} & 1 & 6 & 3000 & consistency \\ \hline
Overall & \multicolumn{1}{c}{16} & 44 & 137 & 8661 & - \\ \hline
\end{tabular}
}
\label{tab:dataset}
\end{table*}

\begin{table*}[]
\centering
\caption{Description of tasks and datasets with the number of human evaluation metrics (\# HE Metrics), number of NLG systems (\# Systems), number of samples (\# Samples), human evaluation aspects (HE Metrics) and System ID for the validation set.}
\resizebox{\textwidth}{20mm}{
\begin{tabular}{ccccccc}
\hline
\textbf{Tasks} & \textbf{Datasets} & \textbf{\# HE Metrics} & \textbf{\# Systems} & \textbf{\# Samples} & \textbf{HE Metrics} & \textbf{System ID} \\ \hline
Data to Text & E2E \cite{dusek-etal-2018-findings} & 1 & 3 & 31 & naturalness & zhang, gong, tnt2 \\ \hline
\multirow{8}{*}{Paraphrase   Generation} & ParaBank1 \cite{hu2019parabank} & 1 & 4 & 69 & overall & 0, 2, 30, 35 \\
 & ParaBank2   \cite{hu2019parabank} & 1 & 4 & 62 & overall & 0, 3, 24, 31 \\
 & ParaBank3   \cite{hu2019parabank} & 1 & 5 & 77 & overall & 4, 0, 6, 30, 35 \\
 & ParaBank4   \cite{hu2019parabank} & 1 & 5 & 84 & overall & 5, 0, 13, 29, 35 \\
 & ParaBank5   \cite{hu2019parabank} & 1 & 5 & 90 & overall & 6, 0, 20, 30, 35 \\
 & ParaBank6   \cite{hu2019parabank} & 1 & 5 & 82 & overall & 7, 0, 6, 29, 35 \\
 & ParaBank7   \cite{hu2019parabank} & 1 & 5 & 77 & overall & 9, 0, 13, 32, 35 \\
 & ParaBank8   \cite{hu2019parabank} & 1 & 5 & 64 & overall & 10, 0, 3, 27, 35 \\ \hline
 Overall  & 9 & 9 & 41 & 636 & - & - \\ \hline
\end{tabular}}
\label{tab:dataset-val}
\end{table*}

\subsection{Learner Selection}\label{ap:learner}
\subsubsection{Practical Recommendation}
We explore the learning stability and accuracy of nine popular statistical machine learning algorithms as the regressors of Learner. 
We replace the core algorithm of Learner in CASF, and carried out experiments in 16 datasets under five NLG tasks. The experiment involved 137 NLG systems and 44 human indicators. The sampling rate is 50\%. The experimental results are shown in the Table \ref{tab:learner}. Each experiment is run three times and the average inter-system ranking accuracy and variance through the three runs of each regressor are recorded. The stability of the regressors and be judged by the recorded variance.

The nine popular statistical machine learning regressors are Linear Regressor \cite{seber2012linear}, AdaBoost \cite{freund1996experiments}, Bootstrap aggregating (Bagging) \cite{breiman1996bagging}, Decision Tree \cite{quinlan1986induction}, Extremely Randomized Trees (ExtRaTree) \cite{geurts2006extremely}, K-Nearest Neighbor (KNN) \cite{cover1967nearest}, Random Forest \cite{breiman2001random}, support vector machine (SVM) \cite{hearst1998support} and Gradient Boosting Decision Tree (GBDT) \cite{friedman2001greedy}. We use the implementation of the corresponding statistical machine learning regressors in the sklearn library.

As for the stability of the algorithm, we do not want the proposed sampling method to get different results in each time of sampling like random sampling. Therefore, we should choose stable regressors as the core algorithm of the learner. The results in Table \ref{tab:learner} show that linear regression, KNN, SVM and GBDT achieve good stability, and the variance of the inter-system ranking of three runs is 0. 
In terms of inter-system ranking accuracy, GBDT obtained the highest inter-system ranking accuracy, reaching 0.83 Kendall's correlation. Based on the above experimental results, we recommend choosing GBDT as the core regression algorithm of the Learner in the proposed CASF.

\subsubsection{Learner Selection in This Paper}
For the selection of regressor for Learner in this paper, we conduct similar experiments on the validation set. Experimental results in Table \ref{tab:learner-val} show 
GBDT obtained the highest inter-system ranking accuracy and stability, reaching 1.000 Kendall's correlation for inter-system ranking accuracy and zero fluctuation. Based on the above experimental results and analysis, we chose GBDT as the core regression algorithm of the Learner in the proposed CASF in this paper.

\begin{table*}[]
\caption{Experiment for practical recommendation of selecting the core regressing method for Learner. 'Mean' represents the average inter-system ranking accuracy across three runs and 'Std' represents the variance of the three runs. HE Metric’ indicates different human evaluation aspects in a dataset. \textbf{Bold number} indicates that the regressor ranks first among all regressors under the corresponding human evaluation metric. \underline{Underlined number} indicates that the regressor ranks second among all regressors for the corresponding human evaluation metric.}
\resizebox{\textwidth}{72mm}{
\begin{tabular}{@{}ccccccccccccccccccccc@{}}
\hline
\multirow{2}{*}{\textbf{Task}} & \multirow{2}{*}{\textbf{Dataset}} & \multirow{2}{*}{\textbf{HE Metric}} & \multicolumn{2}{c}{\textbf{Linear}} & \multicolumn{2}{c}{\textbf{AdaBoost}} & \multicolumn{2}{c}{\textbf{Bagging}} & \multicolumn{2}{c}{\textbf{DecisionTree}} & \multicolumn{2}{c}{\textbf{ExtRaTree}} & \multicolumn{2}{c}{\textbf{KNN}} & \multicolumn{2}{c}{\textbf{Random Forest}} & \multicolumn{2}{c}{\textbf{SVM}} & \multicolumn{2}{c}{\textbf{GBDT}} \\
 &  &  & \textbf{Mean} & \textbf{Std} & \textbf{Mean} & \textbf{Std} & \textbf{Mean} & \textbf{Std} & \textbf{Mean} & \textbf{Std} & \textbf{Mean} & \textbf{Std} & \textbf{Mean} & \textbf{Std} & \textbf{Mean} & \textbf{Std} & \textbf{Mean} & \textbf{Std} & \textbf{Mean} & \textbf{Std} \\ \hline
\multirow{29}{*}{SUM} & \multirow{4}{*}{SummEval} & coherence & 0.617 & 0.000 & 0.628 & 0.150 & 0.583 & 0.191 & 0.622 & 0.136 & 0.672 & 0.198 & 0.650 & 0.000 & 0.522 & 0.142 & 0.717 & 0.000 & 0.950 & 0.000 \\
 &  & consistency & 0.600 & 0.000 & 0.567 & 0.054 & 0.478 & 0.021 & 0.494 & 0.093 & 0.494 & 0.087 & 0.567 & 0.000 & 0.472 & 0.123 & 0.117 & 0.000 & 0.533 & 0.000 \\
 &  & fluency & 0.467 & 0.000 & 0.406 & 0.034 & 0.300 & 0.072 & 0.311 & 0.034 & 0.256 & 0.165 & 0.200 & 0.000 & 0.378 & 0.021 & 0.367 & 0.000 & 0.333 & 0.000 \\
 &  & relevance & 0.750 & 0.000 & 0.472 & 0.122 & 0.450 & 0.072 & 0.611 & 0.244 & 0.461 & 0.162 & 0.817 & 0.000 & 0.567 & 0.167 & 0.383 & 0.000 & 0.817 & 0.000 \\ \cmidrule(l){2-21}
 & REALSumm & litepyramid & 0.399 & 0.000 & 0.430 & 0.109 & 0.394 & 0.021 & 0.333 & 0.119 & 0.403 & 0.054 & 0.601 & 0.000 & 0.529 & 0.041 & 0.464 & 0.000 & 0.543 & 0.000 \\ \cmidrule(l){2-21}
 & \multirow{4}{*}{NeR18} & coherence & 1.000 & 0.000 & 0.841 & 0.224 & 1.000 & 0.000 & 1.000 & 0.000 & 0.968 & 0.045 & 1.000 & 0.000 & 1.000 & 0.000 & 1.000 & 0.000 & 1.000 & 0.000 \\
 &  & fluency & 1.000 & 0.000 & 0.841 & 0.224 & 0.968 & 0.045 & 0.968 & 0.045 & 0.968 & 0.045 & 1.000 & 0.000 & 0.619 & 0.206 & 1.000 & 0.000 & 1.000 & 0.000 \\
 &  & informativeness & 0.714 & 0.000 & 1.000 & 0.000 & 0.905 & 0.135 & 1.000 & 0.000 & 0.873 & 0.119 & 1.000 & 0.000 & 0.873 & 0.119 & 0.714 & 0.000 & 1.000 & 0.000 \\
 &  & relevance & 0.905 & 0.000 & 0.968 & 0.045 & 0.905 & 0.000 & 0.651 & 0.250 & 1.000 & 0.000 & 0.905 & 0.000 & 0.778 & 0.250 & 0.619 & 0.000 & 1.000 & 0.000 \\ \cmidrule(l){2-21}
 & \multirow{4}{*}{DialSummEval} & consistency & 0.513 & 0.000 & 0.632 & 0.074 & 0.726 & 0.032 & 0.752 & 0.024 & 0.675 & 0.151 & 0.872 & 0.000 & 0.547 & 0.044 & 0.615 & 0.000 & 0.769 & 0.000 \\
 &  & relevance & 0.564 & 0.000 & 0.624 & 0.212 & 0.744 & 0.126 & 0.607 & 0.099 & 0.667 & 0.126 & 0.564 & 0.000 & 0.496 & 0.024 & 0.538 & 0.000 & 0.718 & 0.000 \\
 &  & fluency & 0.897 & 0.000 & 0.598 & 0.157 & 0.538 & 0.117 & 0.735 & 0.128 & 0.504 & 0.281 & 0.385 & 0.000 & 0.658 & 0.103 & 0.744 & 0.000 & 0.615 & 0.000 \\
 &  & coherence & 0.795 & 0.000 & 0.684 & 0.067 & 0.590 & 0.091 & 0.786 & 0.067 & 0.632 & 0.154 & 0.641 & 0.000 & 0.556 & 0.053 & 0.846 & 0.000 & 0.897 & 0.000 \\ \cmidrule(l){2-21}
 & \multirow{4}{*}{OpenAI-axis1} & accuracy & 0.000 & 0.000 & 0.400 & 0.432 & 0.400 & 0.283 & 0.267 & 0.377 & 0.333 & 0.340 & 0.000 & 0.000 & 0.400 & 0.432 & 0.200 & 0.000 & 1.000 & 0.000 \\
 &  & coherence & 0.400 & 0.000 & 0.467 & 0.249 & 0.400 & 0.283 & 0.467 & 0.249 & 0.400 & 0.283 & 0.000 & 0.000 & 0.667 & 0.340 & 0.800 & 0.000 & 0.800 & 0.000 \\
 &  & coverage & 1.000 & 0.000 & 0.933 & 0.094 & 1.000 & 0.000 & 1.000 & 0.000 & 0.933 & 0.094 & 1.000 & 0.000 & 1.000 & 0.000 & 0.800 & 0.000 & 0.800 & 0.000 \\
 &  & overall & 1.000 & 0.000 & 0.933 & 0.094 & 0.867 & 0.094 & 1.000 & 0.000 & 0.933 & 0.094 & 1.000 & 0.000 & 0.933 & 0.094 & 0.800 & 0.000 & 1.000 & 0.000 \\ \cmidrule(l){2-21}
 & \multirow{4}{*}{OpenAI-axis2} & accuracy & 0.714 & 0.000 & 0.873 & 0.119 & 0.397 & 0.196 & 0.810 & 0.269 & 0.810 & 0.206 & 0.619 & 0.000 & 0.556 & 0.119 & 0.714 & 0.000 & 0.905 & 0.000 \\
 &  & coherence & 0.905 & 0.000 & 0.270 & 0.119 & 0.524 & 0.156 & 0.587 & 0.314 & 0.492 & 0.273 & 0.238 & 0.000 & 0.397 & 0.119 & 0.429 & 0.000 & 0.429 & 0.000 \\
 &  & coverage & 1.000 & 0.000 & 0.968 & 0.045 & 0.905 & 0.135 & 1.000 & 0.000 & 0.968 & 0.045 & 0.619 & 0.000 & 0.968 & 0.045 & 1.000 & 0.000 & 1.000 & 0.000 \\
 &  & overall & 0.905 & 0.000 & 0.968 & 0.045 & 0.841 & 0.162 & 0.873 & 0.119 & 0.873 & 0.180 & 0.714 & 0.000 & 0.841 & 0.162 & 1.000 & 0.000 & 1.000 & 0.000 \\ \cmidrule(l){2-21}
 & \multirow{4}{*}{OpenAI-CNN/DM1} & accuracy & 0.956 & 0.000 & 0.822 & 0.131 & 0.896 & 0.147 & 0.896 & 0.147 & 0.807 & 0.042 & 0.644 & 0.000 & 0.837 & 0.091 & 0.956 & 0.000 & 0.867 & 0.000 \\
 &  & coherence & 0.822 & 0.000 & 0.556 & 0.181 & 0.407 & 0.137 & 0.837 & 0.171 & 0.407 & 0.302 & 0.600 & 0.000 & 0.333 & 0.063 & 0.244 & 0.000 & 0.600 & 0.000 \\
 &  & coverage & 1.000 & 0.000 & 0.837 & 0.230 & 0.911 & 0.063 & 0.956 & 0.063 & 0.911 & 0.063 & 0.867 & 0.000 & 0.748 & 0.267 & 1.000 & 0.000 & 0.867 & 0.000 \\
 &  & overall & 1.000 & 0.000 & 0.837 & 0.230 & 0.630 & 0.168 & 0.630 & 0.267 & 0.822 & 0.063 & 0.511 & 0.000 & 0.837 & 0.230 & 1.000 & 0.000 & 1.000 & 0.000 \\ \cmidrule(l){2-21}
 & \multirow{4}{*}{OpenAI-CNN/DM3} & accuracy & 1.000 & 0.000 & 1.000 & 0.000 & 0.556 & 0.314 & 0.778 & 0.314 & 0.556 & 0.314 & 1.000 & 0.000 & 0.778 & 0.314 & 1.000 & 0.000 & 1.000 & 0.000 \\
 &  & coherence & 1.000 & 0.000 & 1.000 & 0.000 & 1.000 & 0.000 & 1.000 & 0.000 & 1.000 & 0.000 & 1.000 & 0.000 & 1.000 & 0.000 & 1.000 & 0.000 & 1.000 & 0.000 \\
 &  & coverage & 1.000 & 0.000 & 1.000 & 0.000 & 1.000 & 0.000 & 1.000 & 0.000 & 1.000 & 0.000 & 1.000 & 0.000 & 0.778 & 0.314 & 1.000 & 0.000 & 1.000 & 0.000 \\
 &  & overall & 1.000 & 0.000 & 0.556 & 0.314 & 1.000 & 0.000 & 0.778 & 0.314 & 0.778 & 0.314 & 1.000 & 0.000 & 0.556 & 0.314 & 1.000 & 0.000 & 1.000 & 0.000 \\ \hline
\multirow{6}{*}{MT} & \multirow{2}{*}{\begin{tabular}[c]{@{}c@{}}newstest2020\\       en-de\end{tabular}} & MQM & 0.714 & 0.000 & 0.556 & 0.324 & 0.524 & 0.339 & 0.429 & 0.467 & 0.746 & 0.359 & 0.333 & 0.000 & 0.492 & 0.367 & 1.000 & 0.000 & 0.143 & 0.000 \\
 &  & pSQM & 1.000 & 0.000 & 0.968 & 0.045 & 0.937 & 0.045 & 0.683 & 0.384 & 1.000 & 0.000 & 1.000 & 0.000 & 0.937 & 0.045 & 0.905 & 0.000 & 1.000 & 0.000 \\ \cmidrule(l){2-21}
 & \multirow{2}{*}{\begin{tabular}[c]{@{}c@{}}newstest2020\\       cn-en\end{tabular}} & MQM & 1.000 & 0.000 & 0.619 & 0.243 & 0.714 & 0.404 & 0.667 & 0.269 & 0.476 & 0.332 & 0.286 & 0.000 & 0.452 & 0.388 & 0.214 & 0.000 & 0.929 & 0.000 \\
 &  & pSQM & 0.786 & 0.000 & 0.262 & 0.410 & 0.690 & 0.243 & 0.476 & 0.221 & 0.548 & 0.321 & 0.929 & 0.000 & 0.524 & 0.337 & 0.071 & 0.000 & 0.786 & 0.000 \\ \cmidrule(l){2-21}
 & \begin{tabular}[c]{@{}c@{}}newstest2021\\       cn-en\end{tabular} & MQM & 0.026 & 0.000 & 0.368 & 0.119 & 0.376 & 0.094 & 0.120 & 0.067 & 0.017 & 0.250 & 0.000 & 0.000 & -0.060 & 0.169 & -0.077 & 0.000 & 0.026 & 0.000 \\ \hline
\multirow{6}{*}{DialoGen} & \multirow{6}{*}{Persona Chat} & Understandable & 1.000 & 0.000 & 0.778 & 0.314 & 0.556 & 0.314 & 0.778 & 0.314 & 0.333 & 0.943 & 1.000 & 0.000 & 1.000 & 0.000 & 0.333 & 0.000 & 0.333 & 0.000 \\
 &  & Natural & 1.000 & 0.000 & 0.111 & 0.629 & 0.333 & 0.544 & 0.111 & 0.831 & 1.000 & 0.000 & 1.000 & 0.000 & 0.556 & 0.314 & 0.333 & 0.000 & 1.000 & 0.000 \\
 &  & Maintains Context & 1.000 & 0.000 & 0.333 & 0.943 & 1.000 & 0.000 & 0.333 & 0.943 & 1.000 & 0.000 & 1.000 & 0.000 & 0.333 & 0.943 & 1.000 & 0.000 & 1.000 & 0.000 \\
 &  & Interesting & 1.000 & 0.000 & 1.000 & 0.000 & 0.778 & 0.314 & 1.000 & 0.000 & 1.000 & 0.000 & 1.000 & 0.000 & 1.000 & 0.000 & 1.000 & 0.000 & 1.000 & 0.000 \\
 &  & Uses Knowledge & 1.000 & 0.000 & 1.000 & 0.000 & 1.000 & 0.000 & 1.000 & 0.000 & 1.000 & 0.000 & 1.000 & 0.000 & 1.000 & 0.000 & 1.000 & 0.000 & 1.000 & 0.000 \\
 &  & Overall Quality & 1.000 & 0.000 & 1.000 & 0.000 & 1.000 & 0.000 & 1.000 & 0.000 & 1.000 & 0.000 & 1.000 & 0.000 & 1.000 & 0.000 & 1.000 & 0.000 & 1.000 & 0.000 \\ \hline
\multirow{2}{*}{StoryGen} & MANS-ROC & overall & 1.000 & 0.000 & 1.000 & 0.000 & 1.000 & 0.000 & 1.000 & 0.000 & 1.000 & 0.000 & 1.000 & 0.000 & 1.000 & 0.000 & 1.000 & 0.000 & 1.000 & 0.000 \\ \cmidrule(l){2-21}
 & MANS-WP & overall & 1.000 & 0.000 & 1.000 & 0.000 & 1.000 & 0.000 & 1.000 & 0.000 & 1.000 & 0.000 & 1.000 & 0.000 & 1.000 & 0.000 & 1.000 & 0.000 & 1.000 & 0.000 \\ \hline
\multirow{4}{*}{MMGen} & THUMB-MSCOCO & overall & 1.000 & 0.000 & 1.000 & 0.000 & 1.000 & 0.000 & 1.000 & 0.000 & 1.000 & 0.000 & 1.000 & 0.000 & 1.000 & 0.000 & 1.000 & 0.000 & 1.000 & 0.000 \\ \cmidrule(l){2-21}
& {VATEX-EVAL} & consistency & 1.000 & 0.000 & 0.867 & 0.189 & 0.867 & 0.189 & 0.867 & 0.189 & 0.733 & 0.189 & 1.000 & 0.000 & 1.000 & 0.000 & 1.000 & 0.000 & 1.000 & 0.000 \\ \hline
\multicolumn{3}{c}{Overall Performance} & {\underline 0.828} & 0.000 & 0.727 & 0.158 & 0.729 & 0.126 & 0.732 & 0.171 & 0.738 & 0.150 & 0.740 & 0.000 & 0.701 & 0.154 & 0.724 & 0.000 & \textbf{0.833} & 0.000 \\ \hline
\end{tabular}}
\label{tab:learner}
\end{table*}

\begin{table*}[]
\centering
\caption{Experiment for selecting the core regressing method for Learner on the validation set. 'Mean' represents the average inter-system ranking accuracy across three runs and 'Std' represents the variance of the three runs. HE Metric’ indicates different human evaluation aspects in a dataset. \textbf{Bold number} indicates that the regressor ranks first among all regressors under the corresponding human evaluation metric. \underline{Underlined number} indicates that the regressor ranks second among all regressors for the corresponding human evaluation metric.}
\resizebox{\textwidth}{22mm}{
\begin{tabular}{ccccccccccccccccccccc}
\hline
\multirow{2}{*}{\textbf{Task}} & \multirow{2}{*}{\textbf{Dataset}} & \multirow{2}{*}{\textbf{HE Metric}} & \multicolumn{2}{c}{\textbf{Linear}} & \multicolumn{2}{c}{\textbf{AdaBoost}} & \multicolumn{2}{c}{\textbf{Bagging}} & \multicolumn{2}{c}{\textbf{DecisionTree}} & \multicolumn{2}{c}{\textbf{ExtraTree}} & \multicolumn{2}{c}{\textbf{KNN}} & \multicolumn{2}{c}{\textbf{Random Forest}} & \multicolumn{2}{c}{\textbf{SVM}} & \multicolumn{2}{c}{\textbf{GBDT}} \\ 
 &  &  & \textbf{Mean} & \textbf{Std} & \textbf{Mean} & \textbf{Std} & \textbf{Mean} & \textbf{Std} & \textbf{Mean} & \textbf{Std} & \textbf{Mean} & \textbf{Std} & \textbf{Mean} & \textbf{Std} & \textbf{Mean} & \textbf{Std} & \textbf{Mean} & \textbf{Std} & \textbf{Mean} & \textbf{Std} \\ \hline
Data to Text & E2E & naturalness & 1.000 & 0.000 & 0.556 & 0.629 & 0.556 & 0.629 & -0.111 & 0.314 & 0.333 & 0.544 & 1.000 & 0.000 & 0.111 & 0.629 & 1.000 & 0.000 & 1.000 & 0.000 \\\hline
\multirow{8}{*}{Paraphrase   Generation} & ParaBank1 & overall & 0.667 & 0.000 & 0.333 & 0.471 & 0.333 & 0.272 & 0.333 & 0.471 & 0.444 & 0.416 & 0.667 & 0.000 & 0.889 & 0.157 & 0.000 & 0.000 & 1.000 & 0.000 \\\cmidrule(l){2-21}
 & ParaBank2 & overall & 1.000 & 0.000 & 1.000 & 0.000 & 0.667 & 0.471 & 1.000 & 0.000 & 1.000 & 0.000 & 1.000 & 0.000 & 1.000 & 0.000 & 1.000 & 0.000 & 1.000 & 0.000 \\\cmidrule(l){2-21}
 & ParaBank3 & overall & 1.000 & 0.000 & 1.000 & 0.000 & 1.000 & 0.000 & 1.000 & 0.000 & 1.000 & 0.000 & 1.000 & 0.000 & 0.667 & 0.471 & 0.000 & 0.000 & 1.000 & 0.000 \\\cmidrule(l){2-21}
 & ParaBank4 & overall & 0.667 & 0.000 & 0.556 & 0.416 & 1.000 & 0.000 & 1.000 & 0.000 & 1.000 & 0.000 & 1.000 & 0.000 & 0.889 & 0.157 & 1.000 & 0.000 & 1.000 & 0.000 \\\cmidrule(l){2-21}
 & ParaBank5 & overall & 1.000 & 0.000 & 1.000 & 0.000 & 1.000 & 0.000 & 1.000 & 0.000 & 1.000 & 0.000 & 1.000 & 0.000 & 1.000 & 0.000 & 1.000 & 0.000 & 1.000 & 0.000 \\\cmidrule(l){2-21}
 & ParaBank6 & overall & 1.000 & 0.000 & 1.000 & 0.000 & 0.889 & 0.157 & 0.889 & 0.157 & 1.000 & 0.000 & 1.000 & 0.000 & 0.667 & 0.471 & 1.000 & 0.000 & 1.000 & 0.000 \\\cmidrule(l){2-21}
 & ParaBank7 & overall & 1.000 & 0.000 & 1.000 & 0.000 & 1.000 & 0.000 & 1.000 & 0.000 & 1.000 & 0.000 & 1.000 & 0.000 & 1.000 & 0.000 & 1.000 & 0.000 & 1.000 & 0.000 \\\cmidrule(l){2-21}
 & ParaBank8 & overall & 0.000 & 0.000 & 0.667 & 0.471 & 0.889 & 0.157 & 0.667 & 0.471 & 0.556 & 0.314 & 1.000 & 0.000 & 1.000 & 0.000 & 1.000 & 0.000 & 1.000 & 0.000 \\\hline
\multicolumn{3}{c}{Overall Performance} & 0.815 & 0.000 & 0.790 & 0.221 & 0.815 & 0.187 & 0.753 & 0.157 & 0.815 & 0.142 & {\underline 0.963} & 0.000 & 0.802 & 0.210 & 0.778 & 0.000 & \textbf{1.000} & 0.000 \\ \hline
\end{tabular}}
\label{tab:learner-val}
\end{table*}

\subsection{Automatic Metrics for Preliminary Sampling Phase}\label{ap:auto-matric}
\subsubsection{Practical Recommendation}
By replacing the automated metrics of the proposed CASF in the preliminary sampling phase, we explore which automated metrics are more suitable for measuring sample quality in the preliminary sampling phase. We calculate metrics in the selected NLG automatic metric set by using the official provided code.
Full experiment results of the proposed Constrained Active Sampling Framework on 44 human evaluation metrics from 5 NLG tasks pre-ranking on different automatic metrics are shown in Table \ref{tab:res-metrics-all}.
We can learn from Table \ref{tab:res-metrics-all} that MOVER-SCORE  \cite{zhao2019moverscore} ranks first in the whole inter-system ranking accuracy of 64\% human evaluation metrics. In addition, MOVER-SCORE ranked first in the overall inter-system ranking accuracy of 16 datasets, so we recommend using MOVER-SCORE as the calculation method of sample quality in the preliminary phase.

The results of top-ranked system recognition accuracy shown in Table \ref{tab:res-top1-metrics} demonstrate that MOVER-SCORE has the best recognition performance in summarization, dialogue generation, story generation and multi-modal generation tasks, while the recognition effect in machine translation task is the second-best among 8 automatic metrics. MOVER-SCORE has an average top-ranked system identification accuracy of 93.18\% across all 16 human evaluation metrics, involving 137 NLG systems. These results further indicate that MOVER-SCORE is a more suitable sampling quality measurement method in the preliminary sampling phase of CASF.
\begin{table}[h]
\caption{Experiment results of top-ranked accuracy of the proposed CASF on NLG tasks pre-ranking on different automatic metrics. 
‘Overall’ represents the average result on all human indicators from all tasks.
\textbf{Bold number} indicates that the automatic metric ranks first among all automatic metrics. \underline{Underlined number} indicates that the automatic metric ranks second among all automatic metrics.}
\centering
\resizebox{86mm}{16mm}{
\begin{tabular}{@{}ccccccc@{}}
\hline
Automatic   Metric & SUM & MT & DialoGen & StoryGen & MMGen & Overall \\ \hline
BERT-SCORE & 0.8276 & 0.8000 & 0.8333 & 1.0000 & 1.0000 & 0.8409 \\
MOVER-SCORE & 0.9310 & 0.8000 & 1.0000 & 1.0000 & 1.0000 & \textbf{0.9318} \\
ROUGE-1 & 0.7586 & 1.0000 & 0.8333 & 1.0000 & 1.0000 & 0.8182 \\
ROUGE-2 & 0.7931 & 1.0000 & 0.8333 & 1.0000 & 1.0000 & 0.8409 \\
ROUGE-L & 0.7241 & 1.0000 & 0.8333 & 1.0000 & 1.0000 & 0.7955 \\
BART-SCORE & 0.8621 & 1.0000 & 1.0000 & 0.5000 & 1.0000 & {\underline 0.8864} \\
BLEU & 0.8621 & 1.0000 & 0.8333 & 1.0000 & 1.0000 & {\underline 0.8864} \\
METEOR & 0.7931 & 1.0000 & 0.8333 & 1.0000 & 1.0000 & 0.8409\\ \hline
\end{tabular}}
\label{tab:res-top1-metrics}
\end{table}

\begin{table*}[]
\caption{Experiment results of the proposed Constrained Active Sampling pre-ranking on different automatic metrics. 'HE Metric' indicates different human evaluation aspects in a dataset. \textbf{Bold number} indicates that the automatic metric ranks first among all automatic metrics under the corresponding human evaluation metric. \underline{Underlined number} indicates that the automatic metric ranks second among all automatic metrics for the corresponding human evaluation metric.}
\centering
\resizebox{\textwidth}{80mm}{
\begin{tabular}{@{}ccccccccccc@{}}
\toprule
Task & Dataset & HE Metric & \multicolumn{1}{c}{BERT-SCORE} & \multicolumn{1}{c}{MOVER-SCORE} & \multicolumn{1}{c}{ROUGE-1} & \multicolumn{1}{c}{ROUGE-2} & \multicolumn{1}{c}{ROUGE-L} & \multicolumn{1}{c}{BART-SCORE} & \multicolumn{1}{c}{BLEU} & \multicolumn{1}{c}{METEOR} \\ \midrule
\multirow{29}{*}{Summarization} & \multirow{4}{*}{SummEval} & coherence & 0.5333 & 0.9500 & 0.7333 & 0.7000 & 0.6333 & 0.1667 & 0.8167 & 0.5500 \\
 &  & consistency & -0.0167 & 0.5333 & 0.5833 & 0.3500 & 0.6500 & 0.4833 & 0.2833 & 0.5667 \\
 &  & fluency & 0.2500 & 0.3333 & 0.5833 & 0.2667 & 0.3500 & 0.2500 & 0.0500 & 0.1667 \\
 &  & relevance & 0.6167 & 0.8167 & 0.7000 & 0.3833 & 0.5333 & 0.3833 & 0.6833 & 0.6500 \\  \cmidrule(l){2-11}
 & REALSumm & litepyramid & 0.4928 & 0.5435 & 0.3333 & 0.4493 & 0.5507 & 0.3841 & 0.5217 & 0.4493 \\ \cmidrule(l){2-11}
 & \multirow{4}{*}{NeR18} & coherence & 1.0000 & 1.0000 & 1.0000 & 1.0000 & 1.0000 & 1.0000 & 1.0000 & 1.0000 \\
 &  & fluency & 1.0000 & 1.0000 & 1.0000 & 0.5238 & 1.0000 & 0.9048 & 1.0000 & 1.0000 \\
 &  & informativeness & 0.7143 & 1.0000 & 1.0000 & 1.0000 & 1.0000 & 1.0000 & 1.0000 & 1.0000 \\
 &  & relevance & 1.0000 & 1.0000 & 0.9048 & 0.9048 & 0.9048 & 0.4286 & 1.0000 & 1.0000 \\ \cmidrule(l){2-11}
 & \multirow{4}{*}{DialSummEval} & consistency & 0.7179 & 0.7692 & 0.5128 & 0.6667 & 0.4615 & 0.6923 & 0.6923 & 0.8974 \\
 &  & relevance & 0.5385 & 0.7179 & 0.4359 & 0.6923 & 0.6154 & 0.8718 & 0.4872 & 0.6923 \\
 &  & fluency & 0.6410 & 0.6154 & 0.8718 & 0.5641 & 0.3846 & 0.5385 & 0.5897 & 0.4359 \\
 &  & coherence & 0.6154 & 0.8974 & 0.5897 & 0.5128 & 0.6923 & 0.5128 & 0.3846 & 0.5897 \\ \cmidrule(l){2-11}
 & \multirow{4}{*}{OpenAI-axis1} & accuracy & 0.0000 & 1.0000 & 0.2000 & 0.0000 & 1.0000 & 0.0000 & 0.0000 & 1.0000 \\
 &  & coherence & 0.8000 & 0.8000 & 0.4000 & 0.2000 & 0.0000 & 0.0000 & 0.2000 & 1.0000 \\
 &  & coverage & 1.0000 & 0.8000 & 1.0000 & 1.0000 & 1.0000 & 1.0000 & 1.0000 & 0.8000 \\
 &  & overall & 1.0000 & 1.0000 & 1.0000 & 1.0000 & 1.0000 & 1.0000 & 1.0000 & 0.8000 \\ \cmidrule(l){2-11}
 & \multirow{4}{*}{OpenAI-axis2} & accuracy & 0.6190 & 0.9048 & 0.6190 & 1.0000 & 0.4286 & 0.6190 & 0.7143 & 0.7143 \\
 &  & coherence & 0.7143 & 0.4286 & 0.3333 & 0.6190 & 0.7143 & 0.0476 & 1.0000 & 0.2381 \\
 &  & coverage & 1.0000 & 1.0000 & 1.0000 & 0.9048 & 0.9048 & 0.7143 & 1.0000 & 0.6190 \\
 &  & overall & 1.0000 & 1.0000 & 0.9048 & 1.0000 & 1.0000 & 0.7143 & 1.0000 & 1.0000 \\ \cmidrule(l){2-11}
 & \multirow{4}{*}{OpenAI-CNN/DM1} & accuracy & 0.9556 & 0.8667 & 0.7778 & 1.0000 & 0.7778 & 0.6889 & 0.7778 & 1.0000 \\
 &  & coherence & 0.9556 & 0.6000 & 0.5111 & 1.0000 & 0.6000 & 0.0667 & 0.2889 & 0.5111 \\
 &  & coverage & 0.8667 & 0.8667 & 1.0000 & 1.0000 & 1.0000 & 0.6444 & 0.8667 & 1.0000 \\
 &  & overall & 1.0000 & 1.0000 & 0.8667 & 1.0000 & 1.0000 & 1.0000 & 0.5111 & 1.0000 \\ \cmidrule(l){2-11}
 & \multirow{4}{*}{OpenAI-CNN/DM3} & accuracy & 0.3333 & 1.0000 & 1.0000 & 0.3333 & 0.3333 & 0.3333 & 1.0000 & 1.0000 \\
 &  & coherence & 1.0000 & 1.0000 & 1.0000 & 1.0000 & 1.0000 & 1.0000 & 1.0000 & 1.0000 \\
 &  & coverage & 1.0000 & 1.0000 & 1.0000 & 1.0000 & 1.0000 & 1.0000 & 1.0000 & 1.0000 \\
 &  & overall & 1.0000 & 1.0000 & 0.3333 & 1.0000 & 0.3333 & 1.0000 & 1.0000 & 1.0000 \\ \hline
\multirow{5}{*}{MT} & \multirow{2}{*}{newstest2020 en-de} & MQM & 1.0000 & 0.1429 & 1.0000 & 0.1429 & 0.1429 & 0.3333 & 0.3333 & 0.3333 \\ 
 &  & pSQM & 0.9048 & 1.0000 & 0.9048 & 1.0000 & 0.9048 & 0.9048 & 1.0000 & 0.9048 \\ \cmidrule(l){2-11}
 & \multirow{2}{*}{newstest2020 cn-en} & MQM & 0.7857 & 0.9286 & 0.2143 & 0.7143 & 0.6429 & 0.1429 & 0.2143 & 0.7143 \\
 &  & pSQM & 0.2857 & 0.7857 & 0.5000 & 0.7857 & 0.7857 & 0.2143 & 0.2857 & 0.2143 \\ \cmidrule(l){2-11}
 & newstest2021   cn-en & MQM & -0.0769 & 0.0256 & 0.2308 & 0.1026 & 0.1282 & 0.5897 & 0.0256 & 0.5128 \\ \hline
\multirow{6}{*}{\begin{tabular}[c]{@{}c@{}}Dialogue\\      Generation\end{tabular}} & \multirow{6}{*}{Persona Chat} & Understandable & 1.0000 & 0.3333 & -1.0000 & 0.3333 & 1.0000 & 0.3333 & 0.3333 & 1.0000 \\
 &  & Natural & -0.3333 & 1.0000 & 1.0000 & -0.3333 & 0.3333 & 1.0000 & -1.0000 & 0.3333 \\
 &  & Maintains Context & 1.0000 & 1.0000 & 1.0000 & 1.0000 & 1.0000 & 1.0000 & 1.0000 & 1.0000 \\
 &  & Interesting & 0.3333 & 1.0000 & 1.0000 & 0.3333 & 1.0000 & 1.0000 & 1.0000 & 1.0000 \\
 &  & Uses Knowledge & 1.0000 & 1.0000 & 1.0000 & 1.0000 & 1.0000 & 1.0000 & 1.0000 & 1.0000 \\
 &  & Overall Quality & 1.0000 & 1.0000 & 1.0000 & 1.0000 & 1.0000 & 1.0000 & 1.0000 & 1.0000 \\ \hline
\multirow{2}{*}{Story   Generation} & MANS-ROC & overall & 1.0000 & 1.0000 & 1.0000 & 1.0000 & 1.0000 & 1.0000 & 1.0000 & 1.0000 \\ \cmidrule(l){2-11}
 & MANS-WP & overall & 1.0000 & 1.0000 & -0.4000 & 1.0000 & 1.0000 & 0.8000 & 1.0000 & 1.0000 \\ \hline
\multirow{2}{*}{Multi-Modal   Generation} & THUMB-MSCOCO & overall & 1.0000 & 1.0000 & 1.0000 & 1.0000 & 1.0000 & 1.0000 & 1.0000 & 1.0000 \\  \cmidrule(l){2-11}
 & VATEX-EVAL & overall & 1.0000 & 1.0000 & 1.0000 & 0.6000 & 1.0000 & 0.6000 & 0.6000 & 1.0000 \\ \hline
\multicolumn{3}{c}{Overall} & 0.7329 & \textbf{0.8332} & 0.6965 & 0.6989 & 0.7456 & 0.6446 & 0.6741 & {\underline 0.7885} \\ \bottomrule
\end{tabular}
}
\label{tab:res-metrics-all}
\end{table*}

\begin{table*}[]
\caption{Experiment results of CASF pre-ranking on different automatic metrics on the validation set. 'HE Metric' indicates different human evaluation aspects in a dataset. \textbf{Bold number} indicates that the automatic metric ranks first among all automatic metrics under the corresponding human evaluation metric. \underline{Underlined number} indicates that the automatic metric ranks second among all automatic metrics for the corresponding human evaluation metric.}
\centering
\resizebox{\textwidth}{23mm}{
\begin{tabular}{ccccccccccc}
\hline 
Task & Dataset & HE Metric & \multicolumn{1}{c}{BERT-SCORE} & \multicolumn{1}{c}{MOVER-SCORE} & \multicolumn{1}{c}{ROUGE-1} & \multicolumn{1}{c}{ROUGE-2} & \multicolumn{1}{c}{ROUGE-L} & \multicolumn{1}{c}{BART-SCORE} & \multicolumn{1}{c}{BLEU} & \multicolumn{1}{c}{METEOR} \\\hline
Data to Text & E2E & naturalness & 0.3333 & 1.0000 & 0.3333 & 1.0000 & 1.0000 & -0.3333 & 1.0000 & -0.3333 \\\hline
\multirow{8}{*}{Paraphrase   Generation} & ParaBank1 & overall & 0.0000 & 1.0000 & 1.0000 & 0.6667 & 1.0000 & 1.0000 & 0.0000 & 1.0000 \\\cmidrule(l){2-11}
 & ParaBank2 & overall & 1.0000 & 1.0000 & 1.0000 & 1.0000 & 1.0000 & 1.0000 & 1.0000 & 1.0000 \\\cmidrule(l){2-11}
 & ParaBank3 & overall & 1.0000 & 1.0000 & 1.0000 & 1.0000 & 1.0000 & 1.0000 & 1.0000 & 1.0000 \\\cmidrule(l){2-11}
 & ParaBank4 & overall & 1.0000 & 1.0000 & 1.0000 & 1.0000 & 0.6667 & 1.0000 & 0.6667 & 0.6667 \\\cmidrule(l){2-11}
 & ParaBank5 & overall & 1.0000 & 1.0000 & 1.0000 & 1.0000 & 1.0000 & 1.0000 & 1.0000 & 1.0000 \\\cmidrule(l){2-11}
 & ParaBank6 & overall & 1.0000 & 1.0000 & 1.0000 & 1.0000 & 1.0000 & 1.0000 & 1.0000 & 1.0000 \\\cmidrule(l){2-11}
 & ParaBank7 & overall & 1.0000 & 1.0000 & 1.0000 & 1.0000 & 0.0000 & 1.0000 & 1.0000 & 1.0000 \\\cmidrule(l){2-11}
 & ParaBank8 & overall & 1.0000 & 1.0000 & 1.0000 & 1.0000 & 0.6667 & 0.0000 & 1.0000 & 1.0000 \\\hline
\multicolumn{3}{c}{Overall} & 0.8148 & \textbf{1.0000} & 0.9259 & {\underline 0.9630} & 0.8148 & 0.7407 & 0.8519 & 0.8148 \\ \hline
\end{tabular}}
\label{tab:premetric-val}
\end{table*}

\subsubsection{Automatic Metric Selection in This Paper}
We conduct a similar experiment on the validation set to select automatic metrics for the preliminary phase of CASF in the paper. According to experimental results in Table \ref{tab:premetric-val}, we find MOVER-SCORE is capable to measure sample quality in the preliminary phase. And we finally select MOVER-SCORE as the automatic metric for the preliminary phase of CASF in the paper.

\subsection{Different Sampling Ratio}\label{ap:differentRatio}
Experimental results in Table \ref{tab:diffRatio} show the inter-system ranking accuracy under different sampling ratios. The full results of sampling half of the dataset are in Table \ref{tab:result}. 
Experimental results demonstrate that CASF has the best inter-system ranking accuracy among three different sampling methods under different sampling ratios, with an average gap between random sampling of 0.1133 Kendall correlation while solving the problem of clustered selection and data manipulation for human evaluation.
We also observe an interesting phenomenon that sometimes there is a negative correlation between sampling ratio and inter-system ranking accuracy (with the sampling ratio of 70\% and 80\%), that is, with the increase of sampling ratio, inter-system ranking accuracy decreases. This phenomena may occur because some samples do not contribute to the overall effect or have a negative effect, and are sometimes used as a sign of publication bias \cite{begg1994operating,card2020little}. Overall, the inter-system ranking accuracy increases with the increase of the sampling ratio.

\begin{table*}[]
\caption{Experiment results of Random Sampling, Heuristic Sampling and Constrained Active Sampling Framework (CASF) with different sampling ratio on 16 datasets. The inter-system ranking accuracy recorded in each dataset is the average scores for all aspects of human evaluation under the dataset. \textbf{Bold number} indicates that the sampling method ranks first among all sampling method under the corresponding NLG dataset. Random and Heuristic are performed 3 times and the mean results are recorded.}
\vspace{2mm}
\centering
\resizebox{0.8\textwidth}{84mm}{
\begin{tabular}{@{}ccccccccccc@{}}
\hline
Task & Dataset & Method & 90\% & 80\% & 70\% & 60\% & 40\% & 30\% & 20\% & 10\% \\ \hline
\multirow{24}{*}{SUM} & \multirow{3}{*}{SummEval} & Random & 0.6167 & 0.6236 & 0.5847 & 0.4625 & 0.4306 & 0.3306 & 0.3403 & 0.1097 \\
 &  & Heuristic & 0.6403 & 0.5792 & 0.5403 & \multicolumn{1}{l}{0.5306} & 0.4042 & 0.3611 & 0.3069 & 0.0625 \\
 &  & CASF (ours) & 0.5833 & 0.6417 & 0.5875 & 0.8167 & 0.5000 & 0.4917 & 0.4833 & -0.0708 \\  \cmidrule(l){2-11}
 & \multirow{3}{*}{REALSumm} & Random & 0.6739 & 0.5580 & 0.4928 & 0.5338 & 0.2826 & 0.2874 & 0.3696 & 0.0411 \\
 &  & Heuristic & 0.7657 & 0.6715 & 0.4517 & 0.4324 & 0.2874 & 0.3382 & 0.2923 & 0.0242 \\
 &  & CASF (ours) & 0.9565 & 0.7391 & 0.5797 & 0.5870 & 0.3116 & 0.4275 & 0.4565 & 0.1739 \\ \cmidrule(l){2-11}
 & \multirow{3}{*}{NeR18} & Random & 0.9762 & 1.0000 & 0.9286 & 0.9524 & 0.8492 & 0.6667 & 0.5714 & 0.3810 \\
 &  & Heuristic & 0.9524 & 0.9444 & 0.9762 & 0.8810 & 0.6746 & 0.5635 & 0.3016 & 0.1667 \\
 &  & CASF (ours) & 1.0000 & 1.0000 & 1.0000 & 1.0000 & 0.9762 & 0.9524 & 0.2619 & 0.8095 \\ \cmidrule(l){2-11}
 & \multirow{3}{*}{DialSummEval} & Random & 0.7350 & 0.6966 & 0.6880 & 0.5940 & 0.5919 & 0.4679 & 0.4038 & 0.4423 \\
 &  & Heuristic & 0.7714 & 0.6688 & 0.5641 & 0.6688 & 0.6261 & 0.5278 & 0.4103 & 0.4637 \\
 &  & CASF (ours) & 0.8654 & 0.7308 & 0.6154 & 0.7115 & 0.7372 & 0.7564 & 0.4103 & 0.4872 \\ \cmidrule(l){2-11}
 & \multirow{3}{*}{OpenAI-axis1} & Random & 0.6000 & 0.6833 & 0.7000 & 0.7167 & 0.5500 & 0.7500 & 0.7500 & 0.6000 \\
 &  & Heuristic & 0.7333 & 0.7500 & 0.6500 & 0.7667 & 0.6333 & 0.5500 & 0.4667 & 0.6333 \\
 &  & CASF (ours) & 0.7500 & 0.7500 & 1.0000 & 0.9000 & 0.9500 & 0.9500 & 0.9500 & 0.6500 \\ \cmidrule(l){2-11}
 & \multirow{3}{*}{OpenAI-axis2} & Random & 0.7540 & 0.7540 & 0.7698 & 0.5794 & 0.6270 & 0.5952 & 0.5397 & 0.3492 \\
 &  & Heuristic & 0.8095 & 0.6746 & 0.6746 & 0.6746 & 0.6349 & 0.6905 & 0.4127 & 0.3968 \\
 &  & CASF (ours) & 0.8095 & 0.8571 & 0.9524 & 0.7381 & 0.7381 & 0.9524 & 0.9286 & 0.5000 \\ \cmidrule(l){2-11}
 & \multirow{3}{*}{OpenAI-CNN/DM1} & Random & 0.9667 & 0.8444 & 0.8111 & 0.8259 & 0.5926 & 0.6889 & 0.4741 & 0.6259 \\
 &  & Heuristic & 0.8519 & 0.7593 & 0.7111 & 0.7185 & 0.8185 & 0.6741 & 0.5667 & 0.4444 \\
 &  & CASF (ours) & 0.8778 & 0.8778 & 0.7667 & 0.8444 & 0.8222 & 0.8222 & 0.7556 & 0.7333 \\ \cmidrule(l){2-11}
 & \multirow{3}{*}{OpenAI-CNN/DM3} & Random & 1.0000 & 0.9444 & 0.9444 & 0.9444 & 0.8333 & 0.6667 & 0.5556 & 0.4444 \\
 &  & Heuristic & 0.9444 & 0.8889 & 0.9444 & 0.8889 & 0.8333 & 0.6667 & 0.5000 & 0.2222 \\
 &  & CASF (ours) & 1.0000 & 1.0000 & 1.0000 & 1.0000 & 1.0000 & 0.6667 & 0.6667 & 0.5000 \\ \hline
\multirow{9}{*}{MT} & \multirow{3}{*}{newstext2020 en-de} & Random & 0.8571 & 0.8889 & 0.6984 & 0.7460 & 0.6349 & 0.6032 & 0.5714 & 0.0000 \\
 &  & Heuristic & 0.7302 & 0.7778 & 0.7937 & 0.6032 & 0.4921 & 0.1587 & 0.4921 & 0.1587 \\
 &  & CASF (ours) & 1.0000 & 1.0000 & 1.0000 & 0.6190 & 1.0000 & 0.2381 & 0.6190 & 0.1429 \\ \cmidrule(l){2-11}
 & \multirow{3}{*}{newstext2020 cn-en} & Random & 0.2500 & 0.5952 & 0.3452 & 0.3333 & 0.4167 & 0.1667 & 0.3810 & 0.0000 \\
 &  & Heuristic & 0.7024 & 0.4643 & 0.5000 & 0.4405 & 0.3929 & -0.0595 & 0.2500 & -0.1190 \\
 &  & CASF (ours) & 0.8929 & 0.8929 & 0.8571 & 0.6071 & -0.0855 & 0.2857 & 0.5000 & 0.5000 \\ \cmidrule(l){2-11}
 & \multirow{3}{*}{newstext2021 cn-en} & Random & 0.4103 & 0.2051 & 0.2222 & 0.2564 & -0.0342 & -0.1624 & -0.1880 & -0.1966 \\
 &  & Heuristic & 0.0855 & 0.1282 & 0.1282 & 0.1966 & -0.0342 & -0.1197 & -0.0085 & -0.1624 \\
 &  & CASF (ours) & 0.1026 & 0.0769 & 0.1282 & 0.4615 & 0.0769 & -0.0513 & -0.0513 & -0.1026 \\ \hline
\multirow{3}{*}{DialoGen} & \multirow{3}{*}{Persona Chat} & Random & 0.9259 & 0.7037 & 0.7037 & 0.8148 & 0.4444 & 0.1852 & 0.0741 & -0.0370 \\
 &  & Heuristic & 0.8519 & 0.7778 & 0.8148 & 0.6667 & 0.3333 & 0.4444 & 0.1481 & -0.2593 \\
 &  & CASF (ours) & 1.0000 & 0.8889 & 1.0000 & 0.8889 & 0.6667 & 0.6667 & 0.8889 & 0.4444 \\ \hline
\multirow{6}{*}{StoryGen} & \multirow{3}{*}{MANS-ROC} & Random & 1.0000 & 1.0000 & 1.0000 & 1.0000 & 0.9333 & 1.0000 & 0.6000 & -0.5333 \\
 &  & Heuristic & 1.0000 & 1.0000 & 1.0000 & 1.0000 & 1.0000 & 1.0000 & 0.4000 & -0.3333 \\
 &  & CASF (ours) & 1.0000 & 1.0000 & 1.0000 & 1.0000 & 1.0000 & 1.0000 & 0.8000 & -0.6000 \\ \cmidrule(l){2-11}
 & \multirow{3}{*}{MANS-WP} & Random & 1.0000 & 1.0000 & 1.0000 & 0.9333 & 0.9333 & 1.0000 & 0.5333 & -0.2667 \\
 &  & Heuristic & 1.0000 & 1.0000 & 1.0000 & 1.0000 & 1.0000 & 0.3333 & 1.0000 & -0.3333 \\
 &  & CASF (ours) & 1.0000 & 1.0000 & 1.0000 & 1.0000 & 1.0000 & 1.0000 & 1.0000 & 0.0000 \\ \hline
\multirow{6}{*}{MMGen} & \multirow{3}{*}{THUMB-MSCOCO} & Random & 1.0000 & 1.0000 & 1.0000 & 1.0000 & 0.9333 & 1.0000 & 1.0000 & 0.9333 \\
 &  & Heuristic & 1.0000 & 1.0000 & 1.0000 & 1.0000 & 1.0000 & 1.0000 & 1.0000 & 0.8000 \\
 &  & CASF (ours) & 1.0000 & 1.0000 & 1.0000 & 1.0000 & 1.0000 & 1.0000 & 1.0000 & 1.0000 \\ \cmidrule(l){2-11}
 & \multirow{3}{*}{VATEX-EVAL} & Random & 1.0000 & 1.0000 & 0.8667 & 0.8667 & 0.7333 & 0.5111 & 0.6000 & 0.6444 \\
 &  & Heuristic & 1.0000 & 1.0000 & 0.8667 & 0.8667 & 0.7333 & 0.4667 & 0.5111 & 0.2889 \\
 &  & CASF (ours) & 1.0000 & 1.0000 & 1.0000 & 1.0000 & 0.4667 & 0.6000 & 1.0000 & 1.0000 \\ \hline
\multicolumn{2}{c}{\multirow{3}{*}{Overall   Performance}} & Random & 0.7979 & 0.7811 & 0.7347 & 0.7225 & 0.6095 & 0.5473 & 0.4735 & 0.2211 \\
\multicolumn{2}{c}{} & Heuristic & 0.8024 & 0.7553 & 0.7260 & 0.7084 & 0.6144 & 0.4747 & 0.4406 & 0.1534 \\
\multicolumn{2}{c}{} & CASF (ours) & \textbf{0.8649} & \textbf{0.8409} & \textbf{0.8429} & \textbf{0.8234} & \textbf{0.6975} & \textbf{0.6724} & \textbf{0.6668} & \textbf{0.3855} \\ \hline
\end{tabular}}

\label{tab:diffRatio}
\end{table*}

\subsection{Different Sampling Size}
We treat the sample size as an independent variable and add additional experiments. The experimental results of different sampling sizes are shown in Table \ref{tab:difSS}, and the inter-system ranking accuracy metric is Kendall's Tau. Both Random and Heuristic were run 100 times, and the average inter-system rankings were recorded as Random Mean and Heuristic Mean in Table \ref{tab:difSS}. We also randomly selected three execution results of Random and Heuristic and displayed them in Table \ref{tab:difSS}. Experimental results show that different times of random sampling or heuristic sampling may get different inter-system ranking accuracy. Experimental results also show CASF outperforms the popular NLG human evaluation sampling method Random and Heuristic in typical sampling sizes. We conduct experiments on datasets with a population size larger than the sample size, and the number of tasks(\# Task), datasets(\# Dataset), human evaluation aspects(\# HE Metric), and systems(\# System) involved for each sample size are shown in Table \ref{tab:diffSS-data}. 
\begin{table*}[]
\centering
\caption{Experimental results of different sampling sizes.}
\resizebox{0.65\textwidth}{22mm}{
\begin{tabular}{ccccccc}
\hline
\textbf{Sample Size} & \textbf{50} & \textbf{100} & \textbf{150} & \textbf{200} & \textbf{250} & \textbf{300} \\ \hline
\textbf{Random 1} & 0.6847 & 0.7478 & 0.5938 & 0.7758 & 0.5595 & 0.6639 \\
\textbf{Random 2} & 0.5838 & 0.6648 & 0.7547 & 0.6905 & 0.7105 & 0.6755 \\
\textbf{Random 3} & 0.6012 & 0.7346 & 0.7258 & 0.8062 & 0.6537 & 0.5058 \\
\textbf{Random Mean} & 0.6496 & 0.7478 & 0.7167 & 0.7806 & 0.6596 & 0.6736 \\
\textbf{Heuristic 1} & 0.6460 & 0.7192 & 0.6210 & 0.7768 & 0.5432 & 0.6935 \\
\textbf{Heuristic 2} & 0.5434 & 0.6716 & 0.7542 & 0.7821 & 0.7575 & 0.6112 \\
\textbf{Heuristic 3} & 0.6599 & 0.7490 & 0.7058 & 0.6401 & 0.6435 & 0.5432 \\
\textbf{Heuristic Mean} & 0.6476 & 0.7497 & 0.7137 & 0.7712 & 0.6412 & 0.6725 \\
\textbf{CASF (Ours)} & \textbf{0.7156} & \textbf{0.7514} & \textbf{0.7757} & \textbf{0.8264} & \textbf{0.7706} & \textbf{0.7010} \\ \hline
\end{tabular}}
\label{tab:difSS}
\end{table*}

\begin{table*}[]
\centering
\caption{The number of tasks(\# Task), datasets(\# Dataset), human evaluation aspects(\# HE Metric), and systems(\# System) involved for each sample size.}
\begin{tabular}{ccccccc}
\hline
\textbf{Sample Size} & \textbf{50} & \textbf{100} & \textbf{150} & \textbf{200} & \textbf{250} & \textbf{300} \\ \hline
\textbf{\# Task} & 5 & 4 & 4 & 4 & 3 & 3 \\
\textbf{\# Dataset} & 16 & 14 & 10 & 10 & 6 & 6 \\
\textbf{\# HE Metric} & 44 & 34 & 24 & 24 & 14 & 14 \\
\textbf{\# System} & 137 & 127 & 61 & 61 & 38 & 38 \\ \hline
\end{tabular}
\label{tab:diffSS-data}
\end{table*}

\subsection{Significant Information Retention Accuracy}\label{ap:sig}
We used the common Wilcoxon signed-rank test \cite{demvsar2006statistical} to evaluate the performance of methods on identifying statistically significant differences between systems on the test set. The overall significant information retention accuracy of CASF, Random Sampling and Heuristic Sampling (both iterated 10000 times) in 44 aspects were 0.6030, 0.5992 and 0.5976 at the  $p = 0.05$ level, and 0.4344, 0.4156 and 0.4138 at $p = 0.001$ level when sampling 50\% of the dataset. The results showed CASF outperforms the popular Random Sampling and Heuristic Sampling. 


\subsection{Limitations and Future Work}
Accurate and reliable evaluation of models is an important aspect of NLG research and practical applications. We makes human evaluation more reliable with limited cost and labor used for annotation. However, there are still some limitations. On the one hand, quality of samples are predicted by the Learner with features of automated metrics, which are easy to calculate in practice. The information of automatic indicators may not be comprehensive enough to represent the quality of a sample. Future work would consider introducing the characteristics of samples, such as the length of the generated text and lexical complexity, so as to make the quality of samples more comprehensive. Similarly, future work would take more information into account about redundancy. On the other hand, since reliable human evaluation is important for NLP tasks which are lack of reliable automated metrics, we focuses on the problem of reliable human evaluation in NLG tasks. However, CASF may be applicable to other NLP tasks. We would like to extend CASF to more NLP tasks in future work. 
Due to the necessity of a certain sample size for learner training, our approach may not be applicable in situations with an extremely small sample size, such as when the sample size is less than 50. In cases where the sample size is small, evaluation costs are typically lower, and full-scale assessment could be considered.

\clearpage

\end{document}